\documentclass{article}

\usepackage{arxiv}
\usepackage{times}
\usepackage{epsfig}
\usepackage{graphicx}
\usepackage{amsmath}
\usepackage{amssymb}
\usepackage{booktabs}

\usepackage[breaklinks=true,bookmarks=false]{hyperref}

\newcommand{\argmax}{\mathop{\rm arg~max}\limits}

\newif\ifshowfigs
\showfigstrue 

\newif\ifshowapp
\showapptrue 

\newif\ifshowappdetail
\showappdetailtrue 

\def\bsq#1{
	\lq{#1}\rq
}
\usepackage{amsfonts} 
\usepackage{amsmath} 

\newcommand{\norm}[1]{\left\lVert#1\right\rVert}

\usepackage{mathtools}
\DeclarePairedDelimiterX{\infdivx}[2]{(}{)}{%
	#1\;\delimsize\|\;#2%
}

\usepackage{booktabs} 
\usepackage{graphicx} 
\usepackage{stmaryrd} 
\usepackage{subfigure}
\usepackage{textcomp} 

\usepackage[toc,page,titletoc]{appendix}
\usepackage[capitalise,nameinlink]{cleveref}	

\usepackage{adjustbox}

\usepackage[para]{threeparttable}
\usepackage{array}

\usepackage{color}

\usepackage{dsfont}

\usepackage[thinc]{esdiff}

\usepackage{mathtools}

\usepackage{caption}

\usepackage{float}

\usepackage[bottom]{footmisc}

\usepackage{amsmath}
\usepackage{algorithm}
\usepackage[noend]{algpseudocode}

\algnewcommand{\Initialize}[1]{%
	\State \textbf{Initialize:}
	\State \hspace*{\algorithmicindent}\parbox[t]{0.8\linewidth}{\raggedright #1}
}

\usepackage{authblk} 

\begin{document}

\title{Out-of-Distribution Detection with \\ Reconstruction Error and Typicality-based Penalty}



\author[1,2]{Genki Osada}
\author[1]{Tsubasa Takahashi}
\author[3]{Budrul Ahsan}
\author[2]{Takashi Nishide}

\affil[1]{LINE Corporation, Japan}
\affil[2]{University of Tsukuba, Japan}
\affil[3]{IBM Japan}

\maketitle

\begin{abstract}
 The task of out-of-distribution (OOD) detection is vital to realize safe and reliable operation for real-world applications.
 After the failure of likelihood-based detection in high dimensions had been shown, approaches based on the \emph{typical set} have been attracting attention; however, they still have not achieved satisfactory performance.
 Beginning by presenting the failure case of the typicality-based approach, we propose a new reconstruction error-based approach that employs normalizing flow (NF).
 We further introduce a typicality-based penalty, and by incorporating it into the reconstruction error in NF, we propose a new OOD detection method, penalized reconstruction error (PRE).
 Because the PRE detects test inputs that lie off the in-distribution manifold, it effectively detects adversarial examples as well as OOD examples.
 We show the effectiveness of our method through the evaluation using natural image datasets, CIFAR-10, TinyImageNet, and ILSVRC2012.\footnote{Accepted at WACV 2023.}
\end{abstract}

\section{Introduction}
\label{sec:intro}
Recent works have shown that deep neural network (DNN) models tend to make incorrect predictions with high confidence when the input data at the test time are significantly different from the training data \cite{hendrycks2017a, hendrycks2018deep, liang2018enhancing, NEURIPS2018_abdeb6f5, Yu_2019_ICCV, Hsu_2020_CVPR} or adversely crafted \cite{42503, biggio2013evasion, 43405}.
Such anomalous inputs are often referred to as out-of-distribution (OOD).
We refer to a distribution from which expected data, including training data, comes as the in-distribution (In-Dist) and to a distribution from which unexpected data we should detect comes as the OOD.
We tackle OOD detection \cite{hendrycks2017a, liang2018enhancing, nalisnick2018do, https://doi.org/10.48550/arxiv.2110.11334} which attempts to distinguish whether an input at the test time is from the In-Dist or not.


Earlier, \cite{bishop1994novelty} introduced the likelihood-based approach: detecting data points with low likelihood as OOD using a density estimation model learned on training data.
However, recent experiments using the deep generative models (DGMs) showed that the likelihood-based approach often fails in high dimensions \cite{nalisnick2018do, choi2019generative} (Section \ref{sec:ll}).
This observation has motivated alternative methods \cite{NEURIPS2019_1e795968, Serra2020Input, pmlr-v130-morningstar21a, NEURIPS2020_eddea82a, ijcai2021p292, pmlr-v151-bergamin22a}.
Among them, \cite{choi2019generative, nalisnick2020detecting} argued the need to account for the notion of \emph{typical set} instead of likelihood, but those typicality-based methods still did not achieve satisfactory performance \cite{choi2019generative, pmlr-v139-zhang21g, NEURIPS2020_66121d1f}.
We first argue in Section \ref{sec:tt} the failure case of the typicality-based detection performed on an isotropic Gaussian latent distribution proposed by \cite{choi2019generative, nalisnick2020detecting}, which we refer to as the typicality test in latent space (TTL).
Because the TTL reduces the information in the input vector into a single scalar as the $L_{2}$ norm in latent space, the TTL may lose the information that distinguishes OOD examples from In-Dist ones.

\begin{figure}[t]
	\centering
	\includegraphics[width=0.45 \textwidth]{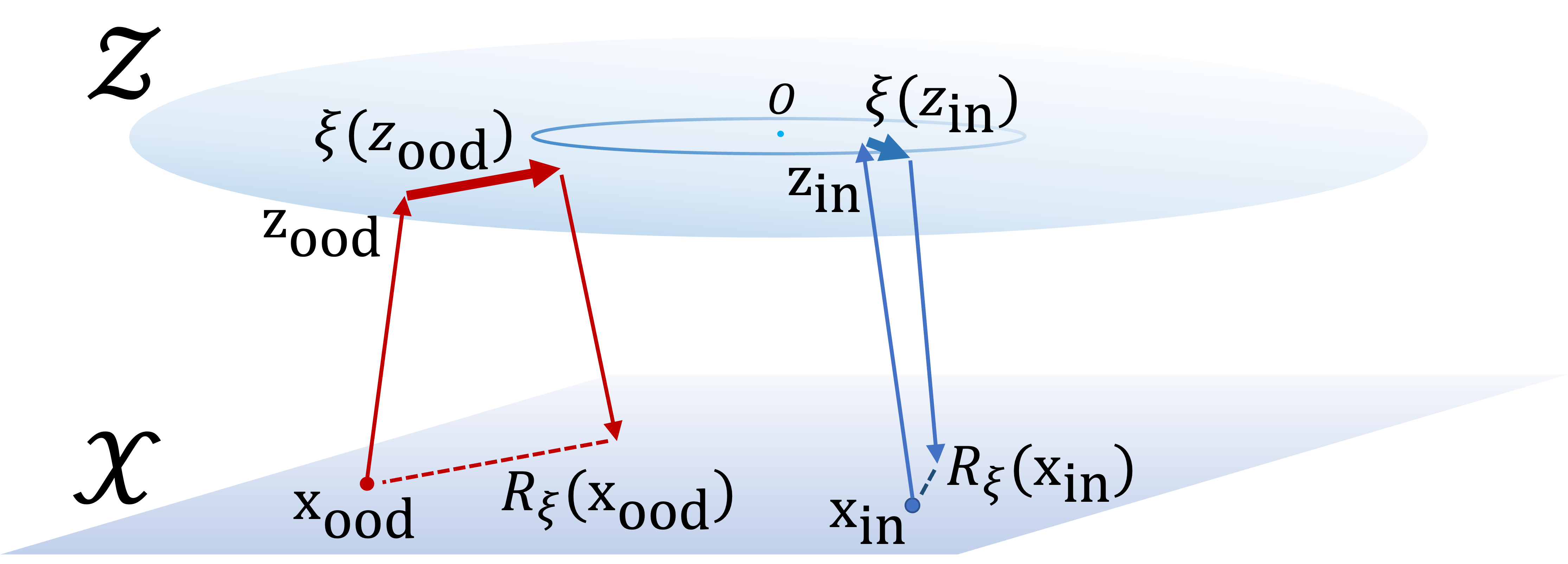}
	\caption{Illustration of our method, PRE. The red dot (${\bf x}_{\text{ood}}$) and blue dot (${\bf x}_{\text{in}}$) represent an OOD and In-Dist samples, respectively. The cyan circle in latent space $\mathcal{Z}$ centered at the origin $O$ represents the Gaussian Annulus, on which the typical set (i.e., In-Dist examples) concentrates. 
		The ${\bf z}_{\text{ood}}$ and ${\bf z}_{\text{in}}$ 
		are subject to controlled perturbations $\xi$ as a penalty (bold arrows), according to $L_{2}$ distance to the Gaussian Annulus.
		While ${\bf z}_{\text{in}}$ will be close to the Gaussian Annulus, ${\bf z}_{\text{ood}}$ will be away from it (Section \ref{sec:tt}), so $|\xi({\bf z}_{\text{ood}})| > |\xi({\bf z}_{\text{in}})|$.
		The reconstruction errors measured in data space $\mathcal{X}$ (the length of the dashed lines) are increased according to $|\xi|$, which leads to $R_{\xi}({\bf x}_{\text{ood}}) > R_{\xi}({\bf x}_{\text{in}})$ and allows us to detect ${\bf x}_{\text{ood}}$.
	}
	\label{fig:overview}
\end{figure}

To address this issue, we first propose a new reconstruction error-based approach that employs normalizing flow (NF)  \cite{pmlr-v37-rezende15, dinh2014nice, dinh2016density}.
We combined the two facts that the previous studies have shown:
1) Assuming the manifold hypothesis  is true \cite{cayton2005algorithms, NIPS2010_3958, fefferman2016testing}, the density estimation model, including NFs, will cause very large Lipschitz constants in the regions that lie off the data manifold \cite{meng2021improved}.
2) The Lipschitz constants of NFs can be connected to its reconstruction error \cite{pmlr-v130-behrmann21a}.
On the premise that the In-Dist examples lie on the manifold yet the OOD examples do not,
we detect a test input that lies off the manifold as OOD when its reconstruction error is large.
Unlike the TTL, our method uses the information of latent vectors as-is, enabling the preservation of the information that distinguishes OOD examples from In-Dist ones.
Second, to boost detection performance further, we introduce a typicality-based penalty.
By applying controlled perturbation (we call a penalty) in latent space according to the \emph{atypicality} of inputs, we can increase the reconstruction error only when inputs are likely to be OOD, thereby improving the detection performance.
The overview of our method is shown in Fig.\ \ref{fig:overview}.

\paragraph{Contribution.}
The contributions of this paper are the following three items:

\setlength{\leftmargini}{0.5cm}
\begin{itemize}

	\item An  OOD detection method based on the reconstruction error in NFs. Based on the property between the Lipschitz constants of NFs and its reconstruction error given by \cite{pmlr-v130-behrmann21a}, the proposed method detects test inputs that lie off the manifold of In-Dist as OOD.

	\item We further introduce a typicality-based penalty. It enhances OOD detection by penalizing inputs atypical in the latent space, on the premise that the In-Dist data are typical. Incorporating this into the reconstruction error, we propose penalized reconstruction error (PRE).

	\item We demonstrate the effectiveness of our PRE in extensive empirical observations on CIFAR-10 \cite{krizhevsky2009learning} and TinyImageNet.
	The PRE consistently showed high detection performance for various OOD types.
	Furthermore, we show on ILSVRC2012 \cite{ILSVRC15} that the proposed methods perform better than $95 \%$ detection in AUROC on average, even for large-size images.
	When an OOD detector is deployed for real-world applications with no control over its input, having no specific weakness is highly desirable.
\end{itemize}

Our PRE also effectively detects adversarial examples.
Among several explanations about the origin of adversarial examples, \cite{tanay2016boundary, jalal2017robust, gilmer2018adversarial, shamir2021dimpled} hypothesized that adversarial examples exist in regions close to, but lie off, the manifold of normal data (i.e., In-Dist data), and \cite{Stutz_2019_CVPR, song2018pixeldefend, samangouei2018defensegan, 10.1145/3133956.3134057}  provided experimental evidence supporting this hypothesis.
Thus, our PRE should also detect adversarial examples as OOD, and we demonstrate it in the evaluation experiments.
Historically, the studies of OOD detection and detecting adversarial examples have progressed separately, and thus few previous works used both samples in their detection performance evaluation, but this work addresses that challenge.

\section{Related Work}

For the likelihood-based methods and typicality test, we refer the reader to Section \ref{sec:ll} and \ref{sec:tt}.
We introduce two other approaches:
%
%
%
1) Reconstruction error
in Auto-Encoder (AE) has been widely used for anomaly detection \cite{Xia_2015_ICCV, 9347460, zong2018deep, Abati_2019_CVPR, NEURIPS2018_5421e013} and also has been employed for detecting adversarial examples \cite{10.1145/3133956.3134057}.
Using an AE model that has been trained to reconstruct normal data (i.e., In-Dist data) well, this approach aims to detect samples that fail to be reconstructed accurately as anomalies (or adversarial examples).
Since the basis of our proposed method is the reconstruction error in NF models, we will evaluate the AE-based reconstruction error method as a baseline.
%
2) Classifier-based methods
that use the outputs of a classifier network has also been taken in many previous works  \cite{hendrycks2017a, hendrycks2018deep, liang2018enhancing, Vyas_2018_ECCV, NEURIPS2018_abdeb6f5, Hsu_2020_CVPR}. 
This approach has also been taken in the works of detecting adversarial examples \cite{feinman2017detecting, FS_ndss18_osada, pmlr-v97-roth19a}.
However, the limitation of this approach is that label information is required for training classifiers.
Furthermore, the dependence on the classifier's performance is their weakness. 
We show that later in the experimental results.


\section{Preliminary}
\subsection{Likelihood-based OOD Detection}
\label{sec:ll}
As an approach to OOD detection,  \cite{bishop1995training} introduced a method that uses a density estimation model learned on In-Dist samples, i.e., training data.
By interpreting the probabilistic density for an input ${\bf x}$, $p({\bf x})$, as a likelihood, it assumes that OOD examples would be assigned a lower likelihood than the In-Dist ones.
Based on this assumption, \cite{song2018pixeldefend} has proposed a method detecting adversarial examples using DGMs, specifically PixelCNN \cite{NIPS2016_b1301141}.
However, \cite{nalisnick2018do, choi2019generative} have presented the counter-evidence against this assumption:
DGMs trained on a particular dataset often assigns higher log-likelihood, $\text{log} \ p({\bf x})$, to OOD examples than the samples from its training dataset (i.e., the In-Dist) in high dimensions.
\cite{choi2019generative, nalisnick2020detecting} argued that this failure of the likelihood-based approach is due to the lack of accounting for the notion of \emph{typical set}:
a set, or a region, that contains the enormous probability mass of distribution
(see \cite{nalisnick2020detecting, cover2012elements} for formal definitions).
Samples drawn from a DGM will come from its typical set; however, in high dimensions, the typical set may not necessarily intersect with high-density regions, i.e., high-likelihood regions.
While it is difficult to formulate the region of the typical set for arbitrary distributions, it is possible for an isotropic Gaussian distribution, which is the latent distribution of NFs  \cite{pmlr-v37-rezende15, dinh2014nice, dinh2016density}. 
It is well known that if a vector ${\bf z}$ belongs to the typical set of the $d$-dimensional Gaussian $\mathcal{N}({\bf \mu}, {\bf \sigma}^{2} \textbf{I}_{\text{d}})$, ${\bf z}$ satisfies $\norm{{\bf z}- {\bf \mu}} \simeq {\bf \sigma} \sqrt{d}$ with a high probability, i.e., concentrates on an annulus centered at ${\bf \mu}$ with radius ${\bf \sigma} \sqrt{d}$, which is known as Gaussian Annulus \cite{vershynin2018high}.
As dimension $d$ increases, the regions on which the typical samples (i.e., In-Dist samples) concentrate move away from the mean of Gaussian, where the likelihood is highest.
That is why In-Dist examples are often assigned low likelihood in high dimensions.

\subsection{Typicality-based OOD Detection}
\label{sec:tt}
\cite{choi2019generative, nalisnick2020detecting} suggested flagging test inputs as OOD when they fall outside of the distribution's typical set.
The deviation from the typical set (i.e., \emph{atypicality}) in a standard Gaussian latent distribution $\mathcal{N}(0, \textbf{I}_{\text{d}})$ of NF is measured as ${\tt abs} (\norm{ {\bf z}} - \sqrt{d})$, where $\norm{ {\bf z}}$ is $L_{2}$ norm of the latent vector corresponding to test input and $ \sqrt{d}$ means the radius of the Gaussian Annulus.
Their proposed method measures the atypicality as an OOD score, which we refer to as the typicality test in latent space (TTL).
The authors however concluded that the TTL was not effective \cite{choi2019generative, pmlr-v139-zhang21g, NEURIPS2020_66121d1f}.
We have the following views regarding the failure of TTL.
As the latent distribution is fixed in NF, In-Dist examples will be highly likely to fall into the typical set, i.e., ${\tt abs} (\norm{ {\bf z}} - \sqrt{d}) \approx 0$.
However, the opposite is not guaranteed: there is no guarantee that OOD examples will always be out of the typical set.
It is because the TTL reduces the information of a vector of test input to a single scalar as its $L_{2}$ norm.
For example, in a 5-dimensional space with the probability density $\mathcal{N}(0, \textbf{I}_{\text{5}})$, the typicality test will judge a vector ${\bf z}$ as In-Dist when $\norm{{\bf z}} \simeq \sqrt{5}$.
At the same time, however, even a vector such as $[\sqrt{5},0,0,0,0]$, which has an extremely low occurrence probability (in the first element) and thereby possibly should be detected as OOD, will judge as In-Dist as well because it belongs to the typical set in terms of its $L_{2}$ norm ($\sqrt{5}$).
Indeed, we observed such an example of this case in the experiments and will show it in Section \ref{sec:tt_fail_analysis}.
We propose a method that addresses this issue.

\subsection{Normalizing Flow}
\label{sec:nf}
The normalizing flow (NF) \cite{pmlr-v37-rezende15, dinh2014nice, dinh2016density} has been becoming a popular method for density estimation.
In short, the NF learns an invertible mapping $f: \mathcal{X} \rightarrow \mathcal{Z}$  that maps the observable data ${\bf x}$ to the latent vector ${\bf z} = f({\bf x})$ where $\mathcal{X} \in \mathbb{R}^{d}$  is a data space and $\mathcal{Z} \in \mathbb{R}^{d}$ is a latent space.
A distribution on $\mathcal{Z}$, which is denoted by $P\text{z}$, is fixed to an isotropic Gaussian $\mathcal{N}(0, \textbf{I}_{\text{d}})$, and thus its density is $p({\bf z}) = (2 \pi)^ {-\frac{d}{2}} \text{exp}(-\frac{1}{2}\norm{{\bf z}}^2)$.
An NF learns an unknown target, or true, distribution on $\mathcal{X}$, which is denoted by $P\text{x}$,  by fitting an approximate model distribution $\widehat{P\text{x}}$ to it.
Under the change of variable rule, the log density of $\widehat{P\text{x}}$  is
$\text{log} \ p({\bf x}) = \text{log} \ p({\bf z}) + \text{log} \ |\text{det} \ J_{f}({\bf x})|$
where $J_{f}({\bf x}) = df({\bf x}) / d{\bf x}$
is the Jacobian matrix of $f$ at ${\bf x}$.
Through maximizing $\text{log} \ p({\bf z})$ and $\text{log} \ |\text{det} \ J_{f}({\bf x})|$ simultaneously w.r.t.\ the samples ${\bf x} \sim P\text{x}$,
$f$ is trained so that $\widehat{P\text{x}}$  matches $P\text{x}$.
In this work, $P\text{x}$ is In-Dist, and we train an NF model using samples from the In-Dist.

	\begin{figure}[t]
	\centering
	\includegraphics[width=0.3 \textwidth]{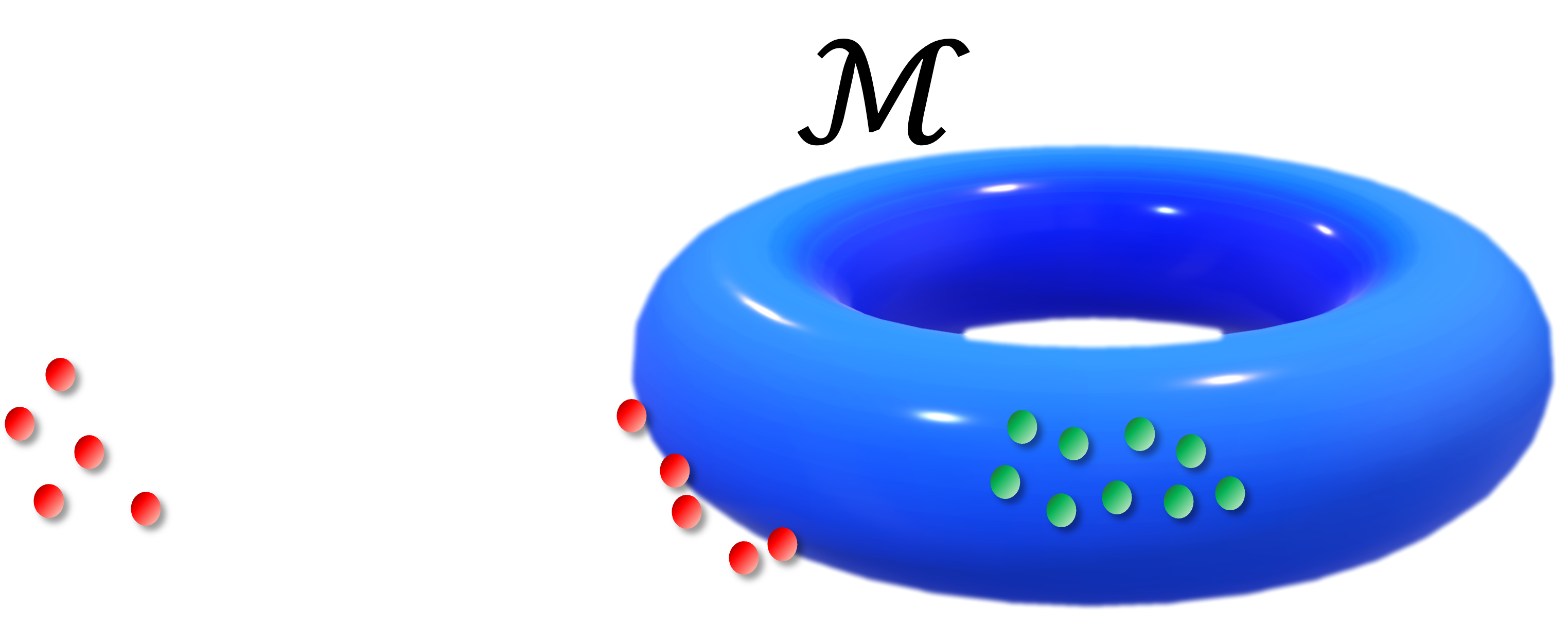}
	\caption{llustration of the manifold $\mathcal{M}$ (represented by torus as an example), In-Dist examples (green dots), and OOD examples close to and far from $\mathcal{M}$ (red dots).}

	\label{fig:manifold}
\end{figure}
\section{Method}
We propose a method that flags a test input as OOD when out of the manifold of In-Dist data (Fig.\ 	\ref{fig:manifold}).
Our method is based on the reconstruction error in NFs, combined with a typicality-based penalty for further performance enhancement, which we call the penalized reconstruction error (PRE)  (Fig.\ \ref{fig:overview}).
Before describing the PRE in Section \ref{sec:method}, 
We first introduce the recently revealed characteristics of NFs that form the basis of the PRE in Section \ref{sec:motivation}: the mechanism of how reconstruction errors occur in NFs, which are supposed to be invertible, and how they become more prominent when inputs are OOD.

\subsection{Motivation to Use Reconstruction Error in NF for OOD Detection}
\label{sec:motivation}
\paragraph{Reconstruction error occurs in NFs.}
Recent studies have shown that when the supports of $P\text{x}$ and $P\text{z}$ are topologically distinct, NFs cannot perfectly fit $\widehat{P\text{x}}$ to $P\text{x}$ \cite{pmlr-v119-cornish20a, pmlr-v119-tanielian20a, verine2021on, DBLP:journals/corr/abs-1903-07714, lu2021implicit}.
As $P\text{z}$ is $\mathcal{N}(0, \textbf{I}_{\text{d}})$, even having \bsq{a hole} in the support of $P\text{x}$ makes them topologically different.
This seems inevitable since $P\text{x}$ is usually very complicated (e.g., a distribution over natural images).
In those works, the discrepancy between $\widehat{P\text{x}}$ and $P\text{x}$ is quantified with the Lipschitz constants.
Let us denote the Lipschitz constant of $f$ by $\text{Lip}(f)$ and that of $f^{-1}$ by $\text{Lip}(f^{-1})$.
When $P\text{x}$ and $P\text{z}$ are topologically distinct,
in order to make $\widehat{P\text{x}}$ fit into $P\text{x}$ well, 
$\text{Lip}(f)$ and $\text{Lip}(f^{-1})$ are required to be significantly large \cite{pmlr-v130-behrmann21a, lu2021implicit}.
Based on this connection, the inequalities to assess the discrepancy between $P\text{x}$ and $\widehat{P\text{x}}$ were presented by \cite{verine2021on} which uses Total Variation (TV) distance and by \cite{pmlr-v119-tanielian20a} which uses the Precision-Recall (PR) \cite{NEURIPS2018_f7696a9b} as well.
\cite{pmlr-v130-behrmann21a} has analyzed it locally from the aspect of numerical errors and introduced another inequality:
\begin{eqnarray}
	\norm{{\bf x} - f^{-1}(f({\bf x}))}  \leq \norm{\text{Lip}(f^{-1})}\norm{\delta_{{\bf z}}}+\norm{\delta_{{\bf x}}}.
	\label{eq:numerical_error}
\end{eqnarray}
See Appendix \ref{sec:lip_and_re} in this paper or  Appendix C in the original paper for the derivation.
The $\delta_{{\bf z}}$ and $\delta_{{\bf x}}$ represent numerical errors in the mapping through $f$ and $f^{-1}$.
When $\text{Lip}(f)$ or $\text{Lip}(f^{-1})$ is large, $\delta_{{\bf z}}$ and $\delta_{{\bf x}}$ grow significantly large and cause ${\tt Inf/NaN}$ values in $f^{-1}(f({\bf x}))$, which is called \emph{inverse explosions}.
Note that Eq.\ \eqref{eq:numerical_error} considers the \emph{local} Lipschitz constant, i.e.,
	$\text{Lip}(f^{-1})^{-1} \norm{{\bf x}_{1}  - {\bf x}_{2}}
	\leq \norm{f({\bf x}_{1}) - f({\bf x}_{2})} \leq \text{Lip}(f) \norm{{\bf x}_{1} - {\bf x}_{2}},
	\forall {\bf x}_{1} , {\bf x}_{2} \in \mathcal{A}$,
	and thus it depends on the region $\mathcal{A}$ where the test inputs exist.
	While computing $\text{Lip}(f^{-1})$, $\delta_{{\bf z}}$, and $\delta_{{\bf x}}$ directly in a complex DNN model is hard \cite{NEURIPS2018_d54e99a6}, Eq.\ \eqref{eq:numerical_error} suggests that the reconstruction error for ${\bf x}$ (the LHS) can approximately measure the discrepancy between $P\text{x}$ and $\widehat{P\text{x}}$ locally.
	Even though the NF is theoretically invertible, \cite{pmlr-v130-behrmann21a} has demonstrated that it is not the case in practice and the reconstruction error is non-zero.
	Another example of the numerical error in an invertible mapping has been observed in \cite{NIPS2017_f9be311e}.

	\paragraph{Connection between OOD and reconstruction errors.}
	\cite{meng2021improved} further connected this discussion to the manifold hypothesis, i.e., that the high-dimensional data in the real world tend to exist on a low-dimensional manifold  \cite{cayton2005algorithms, NIPS2010_3958, fefferman2016testing}.
	Assuming the manifold hypothesis is true, the density $p({\bf x})$ would be very high only if ${\bf x}$ is on the manifold, $\mathcal{M}$, while $p({\bf x})$ would be close to zero otherwise.
	Thus, the value of $p({\bf x})$ may fluctuate abruptly around $\mathcal{M}$.
	It means that the local Lipschitz constants of NFs, i.e., $\text{Lip}(f^{-1})$ in Eq.\ \eqref{eq:numerical_error}, become significantly large, if not infinity.
	As a result, the reconstruction error, which is the lower bound of Eq.\ \eqref{eq:numerical_error}, will be large.
	By contrast, since In-Dist examples should be on $\mathcal{M}$, abrupt fluctuations in $p({\bf x})$ are unlikely to occur.
	Thus $\text{Lip}(f^{-1})$ will be smaller, and the reconstruction error will be smaller for In-Dist examples.
	Thus, this argument allows us to consider an input with a large reconstruction error to be OOD.

	\paragraph{OOD examples close to and far from manifold.}
	In the above, we considered OOD examples ${\bf x}_{\text{ood}}$ that lie off but are close to $\mathcal{M}$ of the In-Dist $P\text{x}$. We depict it as red dots near the blue torus in Fig.\ \ref{fig:manifold}.
	Adversarial examples are considered to be included in such ${\bf x}_{\text{ood}}$.
		On the other hand, there are also ${\bf x}_{\text{ood}}$ far away from $\mathcal{M}$, as depicted by the cluster on the left in Fig.\ 	\ref{fig:manifold}.
		Random noise images may be an example of this.
		In the region far from $\mathcal{M}$, $p({\bf x})$ should be almost constant at 0, so $\text{Lip}(f^{-1})$ may not become large.
		Nevertheless, as we will explain in Section \ref{sec:analysis_l2norm},  the reconstruction error will be large even for such ${\bf x}_{\text{ood}}$ far from $\mathcal{M}$, as long as ${\bf x}_{\text{ood}}$ are regarded as atypical in $P\text{z}$.
		We defer explaining this mechanism to Section \ref{sec:analysis_l2norm}.
		In a nutshell, atypical samples are assigned minimal probabilities in $P\text{z}$, which causes
		$\norm{\delta_{{\bf z}}}$ and $\norm{\delta_{{\bf x}}}$, as opposed to $\text{Lip}(f^{-1})$, to be larger, resulting in a larger reconstruction error.

	\subsection{Our Method: Penalized Reconstruction Error (PRE)}
	\label{sec:method}
	For OOD inputs that lie off the manifold of In-Dist, $\mathcal{M}$, regardless of whether they are close to or far from $\mathcal{M}$, the reconstruction error in NFs will increase.
	Thus we can judge whether a test input is OOD or not by measuring the magnitude of the reconstruction error, written as
	\begin{eqnarray}
		R({\bf x}) := \norm{{\bf x} - f^{-1} \left (f({\bf x}) \right)}.
		\label{eq:re}
	\end{eqnarray}
	Contrary to the PR- and TV-based metrics that need a certain amount of data points to compare $P\text{x}$ and $\widehat{P\text{x}}$,
the reconstruction error works on a single data point, and thus $R({\bf x})$ is suited for use in detection.
	\paragraph{Typicality-based penalty.}
	To further boost detection performance, we add a penalty $\xi$ to a test input ${\bf x}  \in \mathbb{R}^{d}$ in the latent space $\mathcal{Z}$ 
	as $\hat{{\bf z}} = {\bf z} +  \xi$ where ${\bf z} = f({\bf x})  \in \mathbb{R}^{d}$.
	Since an NF model provides a one-to-one mapping between ${\bf x}$ and ${\bf z}$, the shift  by $\xi$ immediately gains the reconstruction error.
	We conducted the controlled experiments to confirm its validity and saw that the degree of the reconstruction error is proportional to the intensity of $\xi$, regardless of the direction of $\xi$.
	(See Appendix \ref{sec:xi_exp}.)
	We want to make $\xi$ large only when ${\bf x}$ is OOD.
	To this end, we use the typicality in $P\text{z}$ described in Section \ref{sec:tt}, and specifically, we design $\xi$ as
	\begin{eqnarray}
		\xi({\bf z})= - \text{sign} \left(\norm{{\bf z}} - \sqrt{d} \right) \left(\frac{\norm{{\bf z}} - \sqrt{d} }{\sqrt{d}}\right)^2.
		\label{eq:xi}
	\end{eqnarray}
	There may be several possible implementations of $\xi({\bf z})$, but we chose to emulate the elastic force in the form of an attractive force proportional to the square of the distance from the center, $\sqrt{d}$.
	The larger the deviation of ${\bf z}$ from the typical set in $P\text{z}$, the larger the value of ${\tt abs} (\norm{ {\bf z}} - \sqrt{d})$ and thus the larger the value of $\xi({\bf z})$.

	\paragraph{Penalized Reconstruction Error (PRE)}
	Incorporating $\xi$, the score we use is:
	\begin{eqnarray}
		R_{\xi}({\bf x}) := \norm{{\bf x} - f^{-1} \left( {\bf z} + \lambda \xi({\bf z}) \frac{{\bf z}}{  \norm{{\bf z}} } \right) }
		\label{eq:rxi}
	\end{eqnarray}
	where 
	$\lambda$ is a coefficient given as a hyperparameter.
	We call the test based on $R_{\xi}$ the penalized reconstruction error (PRE).
	Unlike the TTL that uses the information of ${\bf z} = f({\bf x})$ in a reduced form as $\norm{{\bf z}}$,
	the computation of $R_{\xi}$ uses ${\bf z}$ as-is without reducing.
	Therefore, the PRE works well even for cases where the TTL fails.

	\paragraph{PRE as the OOD detector.}
	The larger $R_{\xi} ({\bf x}_{\text{test}})$, the more likely a test input ${\bf x}_{\text{test}}$ is OOD.
	With using the threshold $\tau$,
	we flags ${\bf x}_{\text{test}}$ as OOD when
	\begin{eqnarray}
		R_{\xi} ({\bf x}_{\text{test}}) > \tau.
	\end{eqnarray}
	The overview of how the PRE identifies the OOD examples is shown in Fig.\ \ref{fig:overview}.
	As the NF model is trained with samples from $P\text{x}$,  if a test input ${\bf x}_{\text{test}}$ belongs to $P\text{x}$ (i.e., ${\bf x}_{\text{test}}$ is In-Dist), ${\bf z}_{\text{test}} = f({\bf x}_{\text{test}})$ would be also typical for $P\text{z}$.
	Then, as $P\text{z}$ is fixed to $\mathcal{N}(0, \textbf{I})$,
	$\norm{{\bf z}_{\text{test}}}$ would be close to $\sqrt{d}$ as described in Section \ref{sec:ll}, and as a result, $\xi({\bf z}_{\text{test}})$ becomes negligible.
	In contrast, if ${\bf x}_{\text{test}}$ is OOD, ${\bf z}_{\text{test}}$ would be atypical for $P\text{z}$, and thus $\norm{{\bf z}_{\text{test}}}$ deviates from $\sqrt{d}$, which makes $\xi({\bf z}_{\text{test}})$ large and consequently enlarges $R_{\xi}({\bf x}_{\text{test}})$.
	Thus, we can view $\xi$ as a deliberate introduction of numerical error $\delta_{{\bf z}}$ in Eq.\ \eqref{eq:numerical_error}.

	We emphasize that the $\xi$ depends only on the typicality of ${\bf x}_{\text{test}}$ and not on its distance from the manifold $\mathcal{M}$.
	Therefore, whenever ${\bf x}_{\text{test}}$ is atypical, $R_{\xi}$ will be large, regardless of whether it is close to or far from  $\mathcal{M}$, i.e., for any OOD data points in Fig.\ \ref{fig:manifold}.


\subsection{Reasons not to employ VAE}

	Variational Auto-Encoder (VAE) \cite{kingma2013auto, pmlr-v32-rezende14} is another generative model that has a Gaussian latent space.
	Although VAE has been used in previous studies of likelihood-based OOD detection, we did not select it for our proposed method 
	for the following reasons:
	1) The primary goal of VAE is to learn a low-dimensional manifold, not to learn invertible transformations as in NF. Therefore, there is no guarantee that the mechanism described in Section \ref{sec:motivation} will hold.
	2) It is known that the reconstructed image of VAE tends to be blurry and that the reconstruction error is large even for In-Dist samples \cite{NEURIPS2018_093f65e0, 10.1007/978-3-030-58452-8_33}; to be used for OOD detection, the reconstruction error must be suppressed for In-Dist samples, and VAE is not suited for this purpose.
	3) Since the latent space of VAE is only an approximation of the Gaussian, $\norm{{\bf z}}$ cannot correctly measure typicality; in previous studies that dealt with both VAE and NF, the typicality test in latent space (TTL) was applied only to NF, for the same reason \cite{choi2019generative, nalisnick2020detecting, pmlr-v130-morningstar21a}.
	In relation to 2) above, we evaluate in Section \ref{sec:exp} the detection performance using the reconstruction error of Auto-Encoder, which is superior to VAE in terms of the small reconstruction error.

\begin{figure*}[t]
	\centering
	\includegraphics[width=0.32 \textwidth]{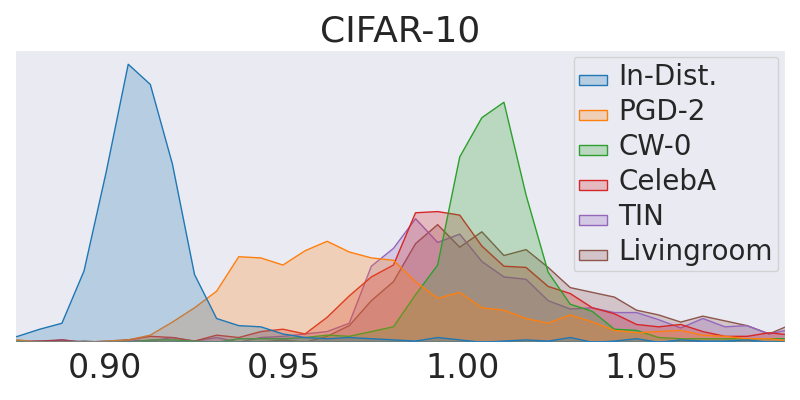}
	\includegraphics[width=0.32 \textwidth]{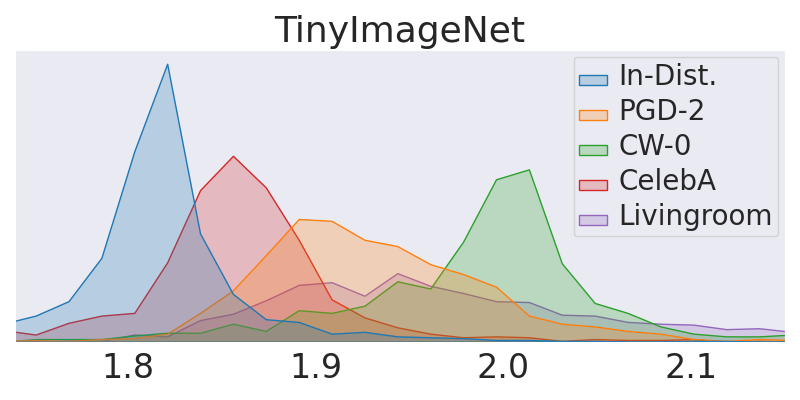}
	\includegraphics[width=0.32 \textwidth]{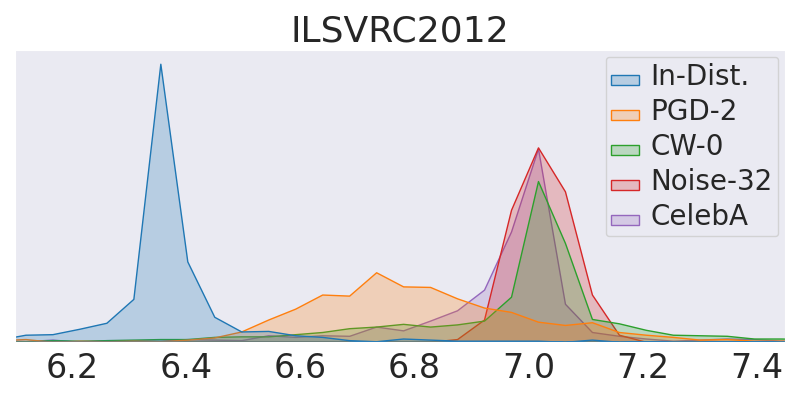}
	\caption{Histograms of the PRE (our method). The x-axis is $R_{\xi}$.
		The $R_{\xi}$ for  \bsq{In-Dist} is lower and well separated than that for OOD datasets. 
	}
	\label{fig:hist_dx_ga}
\end{figure*}
\begin{figure*}[t]
	\centering
	\includegraphics[width=0.32 \textwidth]{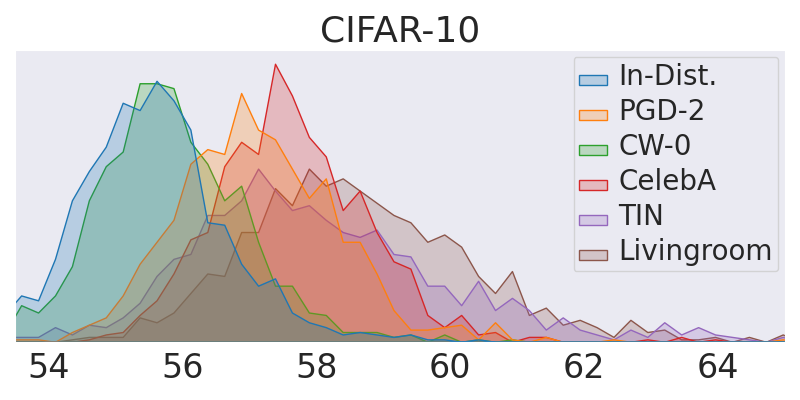}
	\includegraphics[width=0.32 \textwidth]{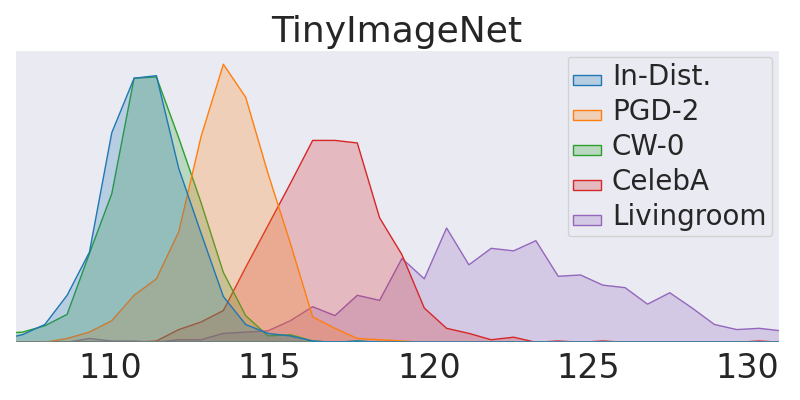}
	\includegraphics[width=0.32 \textwidth]{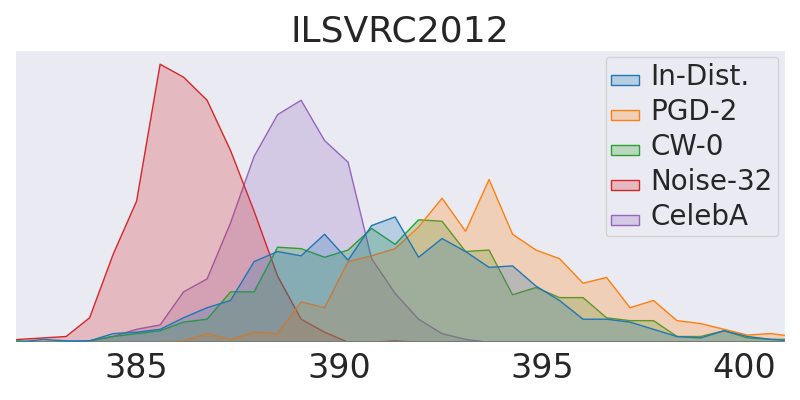}
	\caption{Histograms of $L_{2}$ norm in the latent space. The x-axis is $\norm{{\bf z}}$. The \bsq{In-Dist} is not separate from the OOD datases. In particular, \bsq{CW-0} has almost completely overlapped \bsq{In-Dist}.
	}
	\label{fig:hist_dz}
\end{figure*}
	\section{Experiments}
	\label{sec:exp}
	We demonstrate the effectiveness of our proposed method.
	We measure the success rate of OOD detection using the area under the receiver operating characteristic curve (AUROC) and the area under the precision-recall curve (AUPR).
	The experiments run on a single NVIDIA V100 GPU.


	\subsection{Dataset}

	We utilize three datasets as In-Dist.
	The OOD datasets we use consists of two types: the different datasets from the In-Dist datasets and adversarial examples.
	The tests are performed on 2048 examples that consist of 1024 examples chosen at random from the In-Dist dataset and 1024 examples from the OOD dataset.

	\paragraph{Dataset for In-Dist.}
	We use widely used natural image datasets, CIFAR-10 (C-10), TinyImageNet (TIN), and ILSVRC2012, each of which we call the In-Dist dataset.
	They consist of $32 \times 32$, $64 \times 64$, and $224 \times 224$ pixel RGB images, and the number of containing classes is 10, 200, and 1000, respectively.

	\paragraph{Different datasets from In-Dist.}
	We use CelebA \cite{liu2015faceattributes}, TinyImageNet (TIN) \cite{ILSVRC15}, and LSUN \cite{journals/corr/YuZSSX15} 
	as OOD datasets.
	The LSUN dataset contains several scene categories, from which we chose \emph{Bedroom},  \emph{Living room}, and  \emph{Tower}, and treat each of them as a separate OOD dataset.
	See Appendix \ref{sec:ood_datasets} for processing procedures.
	As ILSVRC2012 contains an extensive range of natural images, including architectural structures such as towers and scenes of rooms, like those included in LSUN datasets,
	LSUN was excluded from the OOD datasets to be evaluated.\footnote{A helpful site for viewing the contained images:  cs.stanford.edu/people/karpathy/cnnembed/}
	Noise images  are also considered one type of OOD.
	Following \cite{Serra2020Input}, we control the noise complexity by varying the size of average-pooling ($\kappa$) to be applied.
	For detailed procedures, refer to Appendix \ref{sec:ood_datasets}.
	Treating images with different $\kappa$ as separate datasets, we refer to them as Noise-$\kappa$.

	\paragraph{Adversarial examples.}
	\label{sec:adv}

	We generate adversarial examples with two methods, Projected Gradient Descent (PGD) \cite{madry2018towards, 45818} and Carlini \& Wagner's (CW) attack \cite{carlini2017towards}.
	Following \cite{pmlr-v80-athalye18a}, we use PGD as $L_{\infty}$ attack and CW as $L_{2}$ attacks, respectively.
	For descriptions of each method, the training settings for the classifiers, and the parameters for generating adversarial examples, we would like to refer the reader to Appendices \ref{sec:classifies} and \ref{sec:adv_appendix}.
	The strength of PGD attacks is controlled by $\epsilon$, which is a parameter often called \emph{attack budgets} that specifies the maximum norm of adversarial perturbation, while the strength of CW attacks is controlled by $k$, called \emph{confidences}.
	We use $\frac{2}{256}$ and $\frac{8}{256}$ for $\epsilon$ in PGD and 0 and 10 for $k$ in CW, which we refer to as PGD-2, PGD-8, CW-0, and CW-10, respectively.
	The classifier we used for both attacks is WideResNet 28-10 (WRN 28-10)\footnote{github.com/tensorflow/models/tree/master/research/autoaugment} \cite{BMVC2016_87} for C-10 and TIN and ResNet-50 v2 \cite{7780459}\footnote{github.com/johnnylu305/ResNet-50-101-152} for ILSVRC2012.
	The classification accuracies before and after attacking are shown in Table \ref{tab:acc_classifier} in Appendix \ref{sec:ood_datasets}.
	For example, on CIFAR-10, 
	the number of samples that could be correctly classified dropped to 57 out of 1024 samples (5.556\%) by the PGD-2 attack,
	and the other three attacks had zero samples successfully classified (0.0\%).

	\subsection{Implementation}
	\paragraph{Normalizing Flow.}
	\label{sec:glow}
	As with most previous works, we use Glow \cite{NIPS2018_8224} for the NF model in our experiments.
	The parameters we used and the training procedure  are described in Appendix \ref{sec:glow_detail}.
	We have experimentally confirmed that the application of data augmentation upon training the Glow is essential for high detection performance.
	We applied the data augmentation of random $2 \times 2$ translation and horizontal flipping on C-10 and TIN.
	For ILSVRC2012, 
	the image is first resized to be $256$ in height or width, whichever is shorter, and cropped to $224 \times 224$ at random.

	\paragraph{Competitors.}
	\label{sec:comparisons}
	We compare the performance of the proposed methods to several existing methods.
	We implement nine competitors:
	the Watanabe-Akaike Information Criterion (WAIC) \cite{choi2019generative},
	the likelihood-ratio test (LLR) \cite{NEURIPS2019_1e795968},
	the Complexity-aware likelihood test (COMP) \cite{Serra2020Input},
	the typicality test in latent space (TTL) \cite{choi2019generative},
	the Maximum Softmax Probability (MSP) \cite{hendrycks2017a},
	the Dropout Uncertainty (DU) \cite{feinman2017detecting},
	the Feature Squeezing (FS) \cite{FS_ndss18_osada},
	the Pairwise Log-Odds (PL) \cite{pmlr-v97-roth19a}, and
	the reconstruction error in Auto-Encoder (AE).
	See Appendix \ref{sec:comparisons_detail} for a description of each method and its implementation.
	For the likelihood-based methods (i.e., WAIC, LLR, and COMP) and TTL, the same Glow model used in our method is used.
	For the classifier-based methods (i.e., MSP, DU, FS, and PL), the classifier is the WRN 28-10 or ResNet-50 v2 which is the same model we used to craft the adversarial examples in the previous section.

	\begin{table*}
	\caption{AUROC (\%) on CIFAR-10. The column labeled as \bsq{Avg.} shows the averaged scores. 
	}
	\label{tab:auroc_C10}
	\centering
	\begin{adjustbox}{max width=\textwidth}
		\begin{threeparttable}[t]
			\vspace{0.20cm}
			\scalebox{0.85}{
				\begin{tabular}{crrrrrrrrrrrr}
					\toprule
					& CelebA    & TIN  & Bed & Living & Tower &  PGD-2  & PGD-8 & CW-0 & CW-10 & Noise-1 & Noise-2 & \underline{Avg.} \\
					\midrule
					WAIC    & 50.36 & 77.94 & 77.25 &84.76  & 79.35 & 41.82 & 73.23 & 47.90 & 46.13 & {\bf  100.0} & {\bf  100.0}&  70.79 \\
					LLR        & 59.77 & 38.05 & 33.33 & 31.42 & 46.01 & 59.10 & 76.09 & 52.28 & 54.36 &  0.80 & 0.61 & 41.07 \\
					COMP  & 72.00 & 82.01 & 78.87 & 87.82  & 5.03 & 60.69 & 98.81 & 51.27 & 53.02 & {\bf  100.0} & {\bf  100.0} & 71.77 \\
					TTL			& 84.87 & 84.19 & 90.39 & 91.36  & 89.68 & 75.22 &  98.99 & 51.14 & 54.15 & {\bf  100.0} & 54.80 & 79.52 \\
					MSP     & 79.32 & 91.00  & 93.83 & 91.67  & 82.05 & 23.85 & 0.0 & {\bf  98.94} & 5.17 & 98.25 & 96.27 &   69.12 \\
					PL         & 81.13 & 63.18 & 54.23  & 49.39  & 59.87 & 78.17 & 97.00 & 56.82 & 80.04 & 24.26 & 	77.07 &  65.56 \\
					FS         & 83.35 & 88.90 & 89.16 & 88.23  & 94.71 & 90.86 & 72.47 & 93.76 & 94.43 &  	91.99  & 96.34 &  89.47 \\
					DU        & 84.64 & 86.33 & 86.53 & 84.76 & 82.04 & 74.62 & 25.54 & 89.33 & 80.84 & 81.09 & 84.61 &  78.21 \\
					AE         & 67.36 & 80.28 & 73.71 & 87.01   & 7.83 & 50.69 & 61.80 & 50.02 & 50.07 & {\bf  100.0}  & 99.52 & 66.21 \\
					\midrule
					RE (ours) & 92.53 & 94.19& 95.92 & 95.83 & 94.35  & 91.66 & 94.58 & 96.08  & 95.09 & 97.45 &  95.80 & 94.86  \\
					PRE (ours) & {\bf 93.62} & {\bf 95.74} & {\bf 97.43 }& {\bf  97.52}& {\bf 95.88 }& {\bf  92.23}&   {\bf   99.93}& 95.00& {\bf 95.21} & {\bf  100.0}  & 96.64 &  {\bf 96.29} \\
					\bottomrule
				\end{tabular}
			}
		\end{threeparttable}
	\end{adjustbox}
\end{table*}

\begin{table*}
	\caption{AUROC (\%) on TinyImageNet. The column labeled as \bsq{Avg.} shows the averaged scores. 
	}
	\label{tab:auroc_TIN}
	\centering
	\begin{adjustbox}{max width=\textwidth}
		\begin{threeparttable}[t]
			\vspace{0.20cm}
			\scalebox{0.85}{
				\begin{tabular}{crrrrrrrrrrr}
					\toprule
					& CelebA     & Bed & Living & Tower  &  PGD-2  & PGD-8 & CW-0 & CW-10 & Noise-1 & Noise-2&  \underline{Avg.} \\
					\midrule
					WAIC    & 11.92 & 63.54 & 67.75 & 72.87  & 40.49 & 49.78 & 48.44 & 46.05 & {\bf 100.0} & {\bf 100.0}  &  60.08  \\
					LLR       & 92.76 & 69.95 & 70.19 & 78.27 & 58.45 & 96.78 & 51.66 & 53.86 & 50.57& 0.0 & 62.25 \\
					COMP  & 39.83 & 48.02 & 61.61 & 46.72  & 55.98 & 95.01 & 50.54 & 52.11 & {\bf 100.0} & {\bf 100.0}  &  64.98 \\
					TTL       &  {\bf 97.51} & 98.78 & 99.36 & {\bf 98.68}  & 83.47 & {\bf 100.0} & 51.89 & 57.73 & {\bf 100.0} & {\bf 100.0} & 88.74 \\
					MSP     & 76.88 & 77.51 & 73.38 & 73.91  & 6.91 & 0.0 & 69.09 & 4.20 & 68.34 & 74.76 &  52.50 \\
					PL         & 49.74 & 27.08 & 26.26 & 28.22 & 94.24 & 99.97 & 34.76 & 87.66 & 15.04 & 31.63 &  49.46 \\
					FS         & 29.21 & 27.26 & 29.36 & 25.70 & 71.98 & 16.86 & 50.64 & 82.21 & 42.17 & 50.27 &  42.57 \\
					DU        & 47.06 & 37.12 & 32.52 & 27.98 & 50.94 & 68.21 & 49.02 & 50.83 & 48.99 & 22.47 & 43.51 \\
					AE         & 14.92 & 20.90 & 32.94 & 23.74 & 49.89 & 52.12 & 50.00 & 50.03 & 95.45& 70.37 & 46.04 \\
					\midrule
					RE (ours)   & 46.68 & 61.97 & 62.26 & 59.51 & 92.86 & 92.94 & {\bf  92.43} & 93.26  &  98.53 & 98.45 & 79.89  \\
					PRE (ours) & 95.55& {\bf 99.01} & {\bf 99.46} & 97.37  & {\bf 95.04} & {\bf 100.0} & 92.42 & {\bf 94.93} & {\bf 100.0}  &{\bf 100.0} &  {\bf 97.38} \\
					\bottomrule
				\end{tabular}
			}
		\end{threeparttable}
	\end{adjustbox}
\end{table*}

	\subsection{Results}
	We measure the success rate of OOD detection using the area under the receiver operating characteristic curve (AUROC) and the area under the precision-recall curve (AUPR).
	The higher is better for both.
	The results in AUPR are presented in Appendix \ref{sec:aupr}.
	We denote the reconstruction errors in NFs without the penalty $\xi$ by RE.
	We also show the histograms in Fig.\ \ref{fig:hist_dx_ga}  for PRE (i.e., $R_{\xi}$) and Fig.\ \ref{fig:hist_dx}  (in Appendix \ref{sec:dx}) for RE (i.e., $R$).
	As for $\lambda$ in Eq.\ \eqref{eq:rxi}, we empirically chose $\lambda=50$ for CIFAR-10 and $\lambda=100$ for TinyImageNet and ILSVRC2012. 
	(The performance with different $\lambda$ is presented in Appendix \ref{sec:lambda}.)

	\paragraph{CIFAR-10 and TinyImageNet.}
	Tables \ref{tab:auroc_C10} and \ref{tab:auroc_TIN} show AUROC for C-10 and TIN.
	The PRE performed best for the majority of OOD datasets (the best scores are shown in bold).
	Importantly, unlike the other methods, the PRE exhibited high performance over all the cases:
	the columns of Avg.\ show that the PRE significantly outperformed the existing methods in average scores.
	When the detection method is deployed for real-world applications with no control over their input data, having no specific weakness is a strong advantage of PRE.
	The RE  showed the second-best performance after the PRE on C-10.
	On TIN, while the RE performed well in the cases where the TTL failed (i.e., CW, which we discuss in Section \ref{sec:tt_fail_analysis}),
	the RE performed poorly in CelebA, Bed, Living, and Tower.
	Notably, however, the penalty $\xi$ proved to be remarkably effective in those cases, and the PRE combined with $\xi$ improved significantly.
	From the comparison of PRE and RE in the tables, we see that the performance is improved by $\xi$.
	It is shown that the performance of likelihood-based methods (WAIC, LLR, and COMP)  is greatly dependent on the type of OOD.
	On C-10, the performance of classifier-based methods 
	is relatively better than the likelihood-based methods,
	however, their performance was significantly degraded on TIN.
	In accordance with the increase in the number of classification classes from 10 (in C-10) to 200 (in TIN), the classifier's performance decreased, which caused the detection performance to decrease.
	It may be the weakness specific to the classifier-based methods.
	The performance of AE was at the bottom.
	Similar results were observed in AUPR as well.

	\paragraph{ILSVRC2012.}
	Table \ref{tab:auroc_IN} shows the AUROC for ILSVRC2012.
	It suggests that the proposed methods perform well even for large-size images.
	We found that the reconstruction error alone (i.e., RE) could achieve high performance and the effect of the penalty was marginal on ILSVRC2012.

	\begin{table*}[t]
		\caption{AUROC (\%) on ILSVRC2012 with our methods. The column labeled as \bsq{Avg.} shows the averaged scores.}
		\label{tab:auroc_IN}
		\centering
		\begin{adjustbox}{max width=\textwidth}
			\begin{threeparttable}[t]
				\vspace{0.20cm}
				\scalebox{0.85}{
					\begin{tabular}{crrrrrrrr}
						\toprule
						& CelebA     & PGD-2     & PGD-8 & CW-0 & CW-10 & Noise-2 & Noise-32 &  \underline{Avg.} \\
						\midrule
						RE    & 94.65 & 93.96 & 96.42 & {\bf 94.59} & 94.87 & 97.24 & {\bf 97.68} &  95.63 \\
						PRE & {\bf  94.89} & {\bf  94.24} & {\bf  96.66} & 94.58 & {\bf 95.64} & {\bf 97.75} & {\bf 97.68} & {\bf 95.92}\\
						\bottomrule
					\end{tabular}
				}
			\end{threeparttable}
		\end{adjustbox}
	\end{table*}


	\section{Discussion}

	\subsection{Analysis with Tail Bound}
	\label{sec:analysis_l2norm}
	In Section \ref{sec:motivation} we explained how $R$ (and $R_{\xi}$) increase for OOD examples ${\bf x}_{\text{ood}}$ close to the manifold $\mathcal{M}$.
	This section discusses how the proposed methods detect ${\bf x}_{\text{ood}}$ far from $\mathcal{M}$, using the tail bound.
	\paragraph{OOD examples are assigned minimal probabilities in latent space.}
	The intensity of the penalty for a particular input ${\bf x}$, $\xi(f({\bf x}))$,  depends on how its $L_{2}$ norm in the latent space $\mathcal{Z} \in \mathbb{R}^{d}$ (i.e.,$\norm{{\bf z}} = \norm{f({\bf x})}$) deviates from $\sqrt{d}$, as Eq.\ \eqref{eq:rxi}.
	 We show the histograms for $\norm{{\bf z}}$ in Fig.\ \ref{fig:hist_dz}. 
	We note that $\sqrt{d}$ is about $55.43$ for C-10, $110.85$ for TIN, and $387.98$ for ILSVRC2012.
	Thus, we see that the distribution modes for In-Dist examples are consistent approximately with the theoretical value, $\sqrt{d}$ (though it is slightly biased toward larger values on ILSVRC2012).
	At the same time, we see that $\norm{{\bf z}}$ for ${\bf x}_{\text{ood}}$ deviate from $\sqrt{d}$, except for CW's examples.
	We assess the degree of this deviation with the Chernoff tail bound for the $L_{2}$ norm of i.i.d. standard Gaussian vector ${\bf z} \in \mathbb{R}^{d}$:
	for any $\epsilon \in (0,1)$ we have
\begin{eqnarray}
	\text{Pr} \left[ d(1 - \epsilon) < \norm{{\bf z}}^{2}  < d(1 + \epsilon) \right]  \geq  1 - 2 \text{exp} \left( - \frac{d \epsilon^{2}}{8}\right).
	\label{eq:tailbound}
\end{eqnarray}
See Appendix \ref{sec:tailbound} for the derivation of Eq.\ \eqref{eq:tailbound} and the analysis described below.
When $d = 3072$ (i.e., on C-10) and we set $\epsilon = 0.32356413$, we have $\text{Pr} \left[ \norm{{\bf z}} > 63.765108 \right] \leq \frac{1 }{ 2^{58}} $,  for instance.
It tells us that the probability that a vector ${\bf z} \in \mathbb{R}^{3072}$ with its $L_{2}$ norm $63.87$, which corresponds to the median value of $\norm{{\bf z}}$ over 1024 PGD-8 examples on C-10 we used, occurs in $P\text{z}$ (i.e., $\mathcal{N}(0, \textbf{I}_{3072})$) is less than $\frac{1 }{ 2^{58}} = 3.4694 \mathrm{e}{-18}$.
As another example, the median value of $\norm{{\bf z}}$ for 1024 CelebA (OOD) examples in $\mathcal{Z}$ built with TIN (In-Dist) is $116.81$.
The probability of observing a vector ${\bf z} \in \mathbb{R}^{12288}$  sampled from $\mathcal{N}(0, \textbf{I}_{12288})$ with its $L_{2}$ norm $116.81$ is less than $\frac{1 }{ 2^{26}} = 1.4901\mathrm{e}{-08}$, as Eq. \eqref{eq:tailbound} gives $\text{Pr}  \left[ \norm{{\bf z}} > 116.700553 \right] \leq \frac{1 }{ 2^{26}}$ with $\epsilon =0.108318603$.
As such, the tail bound shows that those (OOD) examples are regarded as extremely rare events in $P\text{z}$ built with the In-Dist training data.

\paragraph{Small probabilities increase the reconstruction error regardless of the distance to the manifold.}
The above observation implies the following:
(OOD) examples not included in the typical set of $P\text{z}$ are assigned extremely small probability.
 It leads to a decrease in the number of significant digits of the probabilistic density $p(\text{z})$ in the transformation of the NF, and it may cause the rounding error of floating-point.
Those rounding errors increase $\delta_{{\bf x}}$ and $\delta_{{\bf z}}$ in Eq.\ \eqref{eq:numerical_error}, increasing the reconstruction error in NFs, $R$ (and hence $R_{\xi}$ that the PRE uses).
In other words, it suggests that atypical examples in $P\text{z}$ will have larger $R$ (and $R_{\xi}$), independent of the distance to $\mathcal{M}$.
The Noise-$\kappa$ dataset samples used in our experiments (Section \ref{sec:exp}) are possibly far from $\mathcal{M}$.
Both PRE and RE showed high detection performance even against Noise-$\kappa$ datasets.
We understand that the mechanism described here is behind this success.


\subsection{Analysis of Typicality Test Failures}
\label{sec:tt_fail_analysis}
Lastly, we analyze why the typicality test (TTL) failed to detect CW’s adversarial examples (and some Noise datasets)
(Tables \ref{tab:auroc_C10} and \ref{tab:auroc_TIN}).
We addressed it by arbitrarily partitioning latent vectors ${\bf z} \in \mathbb{R}^{3072}$ on C-10 
into two parts as [${\bf z}_{a}, {\bf z}_{b}] = {\bf z}$  where ${\bf z}_{a} \in \mathbb{R}^{2688}$ and ${\bf z}_{b} \in \mathbb{R}^{384}$, and measuring the $L_{2}$ norm separately for each.
(Due to space limitations, see Appendix \ref{sec:za_zb} for details of the experiment.)
We then found that the deviation from the In-Dist observed in ${\bf z}_{a}$ and ${\bf z}_{b}$ cancels out, and as a result, $\norm{{\bf z}}$ of CW's examples becomes indistinguishable from those of In-Dist ones.
This is exactly the case described in Section \ref{sec:tt} where the TTL fails.
The typicality-based penalty in the PRE is therefore ineffective in these cases.
However, in the calculation of the reconstruction error (RE), which is the basis of the PRE, the information of ${\bf z}$ is used as-is without being reduced to the $L_{2}$ norm, and the information on the deviation from the In-Dist in each dimension is preserved.
Consequently, it enables a clear separation between In-Dist and OOD

\section{Limitation}
\label{sec:limit}
1) The penalty $\xi$ we introduced is by design ineffective for OOD examples for which the TTL is ineffective.
However, we experimentally show that it improves performance in most cases.
2) The adversarial examples have been shown to be  generated off the data manifold \cite{Stutz_2019_CVPR};
however, it has also been shown that it is possible to generate the ones lying on the manifold deliberately\cite{Stutz_2019_CVPR, jalal2017robust, gilmer2018adversarial}.
Since the detection target of our method is off-manifold inputs, we left such on-manifold examples out of the scope in this work.
If it is possible to generate examples that are on-manifold and at the same time typical in the latent space,
maybe by \bsq{adaptive attacks} \cite{pmlr-v80-athalye18a},
it would be difficult to detect them with the PRE.
This discussion is not limited to adversarial examples but can be extended to OOD in general, and we leave it for future work.

\section{Conclusion}
We have presented PRE,  a novel method that detects OOD inputs lying off the manifold.
As the reconstruction error in NFs increases regardless of whether OOD inputs are close to or far from the manifold, PRE can detect them by measuring the magnitude of the reconstruction error.
We further proposed a technique that penalizes the atypical inputs in the latent space to enhance detection performance.
We demonstrated state-of-the-art performance with PRE on CIFAR-10 and TinyImageNet and showed it works even on the large size images, ILSVRC2012.

\bibliographystyle{unsrt} 
\bibliography{my_bib_220808}

\clearpage

\appendix

\section{Connection between Lipschitz Constant and Reconstruction Error}
\label{sec:lip_and_re}
For the sake of completeness, we present the derivation of Eq.\ \eqref{eq:numerical_error} from Appendix C in \cite{pmlr-v130-behrmann21a}.
We consider the following two forward mappings:
\begin{eqnarray*}
	{\bf z} &=& f({\bf x}), \\
	{\bf z}_{\delta} &=&  f_{\delta}({\bf x}) := {\bf z} + \delta_{{\bf z}},
\end{eqnarray*}
where $f$ is an analytically exact computation, whereas $f_{\delta}$ is a practical floating-point inexact computation.
Similarly, we consider  the following two inverse mappings:
\begin{eqnarray*}
	{\bf x}_{\delta_{1}}  &=& f^{-1}({\bf z}_{\delta}), \\
	{\bf x}_{\delta_{2}} &=&  f^{-1}_{\delta}({\bf z}_{\delta}) := {\bf x}_{\delta_{1}} + \delta_{{\bf x}}.
\end{eqnarray*}
The $\delta_{{\bf z}}$ and $\delta_{{\bf x}}$ are numerical errors in the mapping through $f_{\delta}$ and $f^{-1}_{\delta}$.
By the definition of the Lipschitz continuous, we obtain the following:
\begin{eqnarray*}
	\frac{ \norm{f^{-1}({\bf z}) - f^{-1}({\bf z}_{\delta})} }{\norm{ {\bf z} - {\bf z}_{\delta} }} = \frac{ \norm{ {\bf x} - {\bf x}_{\delta_{1}} }}{\norm{ {\bf z} - {\bf z}_{\delta} }}  \leq \text{Lip}(f^{-1}) \\
	\Leftrightarrow \norm{ {\bf x} - {\bf x}_{\delta_{1}} } = \text{Lip}(f^{-1}) \norm{ {\bf z} - {\bf z}_{\delta} } = \text{Lip}(f^{-1}) \norm{ \delta_{{\bf z}}}.
\end{eqnarray*}
We now consider the reconstruction error with $f_{\delta}$ and $f^{-1}_{\delta}$:
\begin{eqnarray*}
	\norm{ {\bf x} -  f^{-1}_{\delta}( f_{\delta}({\bf x} )   )} &=&  \norm{ {\bf x} - {\bf x}_{\delta_{2}} } \\
	&=&  \norm{ {\bf x}  - ({\bf x}_{\delta_{1}} + \delta_{{\bf x}})} \\
	& \leq & \norm{ {\bf x}  - {\bf x}_{\delta_{1}}  }  + \norm{  \delta_{{\bf x}}} \\
	& = &  \text{Lip}(f^{-1}) \norm{ \delta_{{\bf z}}} + \norm{  \delta_{{\bf x}}}.
\end{eqnarray*}
While the reconstruction error is zero in the ideal case using $f$ and $f^{-1}$, the upper bound on the reconstruction error is given above when we consider $f_{\delta}$ and $f^{-1}_{\delta}$ to account for the numeral error caused by the large Lipschitz coefficients in practice.

\section{Controlled Experiments of Penalty in Latent Space.}
\label{sec:xi_exp}

\begin{figure}[h]
	\centering
	\includegraphics[width=0.23 \textwidth]{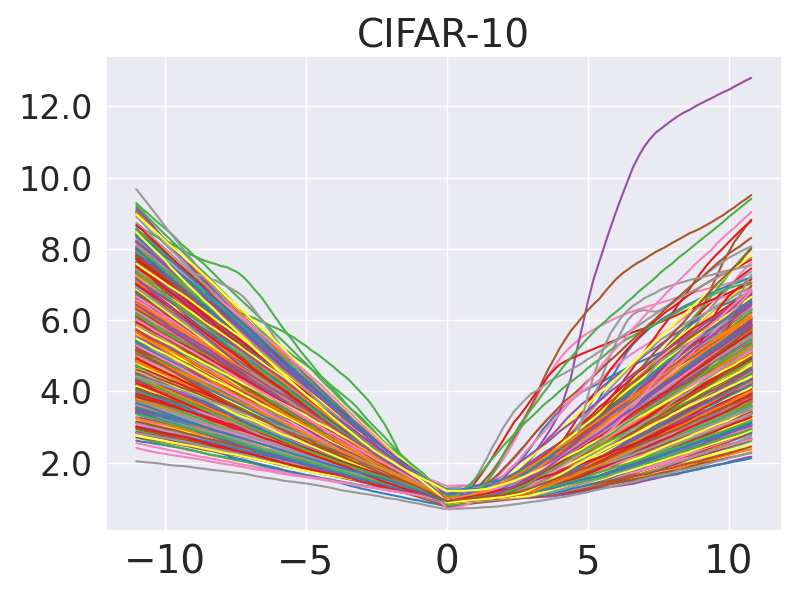}
	\includegraphics[width=0.23 \textwidth]{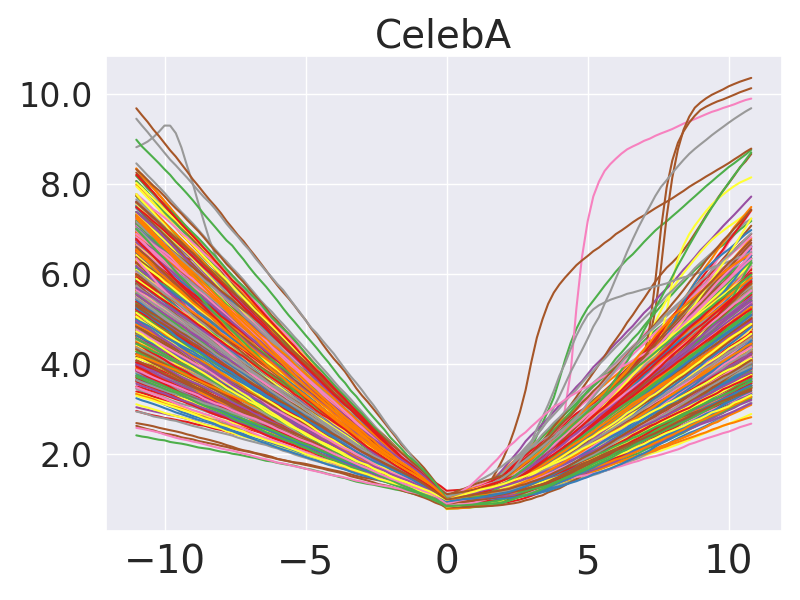}
	\caption{The effects of penalty in latent space. The x-axis is the intensity of penalty ($\xi'$) and the y-axis is penalized reconstruction error ($R'$). Each line corresponds to each example. For both figures, In-Dist samples used to train the Glow model are from CIFAR-10. The OOD samples are from CIFAR-10's test dataset on the left figure and are from CelebA on the right figure.
	}
	\label{fig:xi_exp}
\end{figure}

We show experimentally that penalty $\xi$  in the latent space increases the reconstruction error.
Letting $R'({\bf x}) = \norm{{\bf x} - f^{-1}(\hat{{\bf z}}))}$ and $\hat{{\bf z}} = {\bf z} + \xi' \frac{{\bf z}}{  \norm{{\bf z}} } $ with ${\bf z} = f({\bf x})$,
we measured $R'({\bf x})$ for synthetically generated $\xi'$ in the range of $(-11,11)$ with $0.2$ increments.
When $\xi' > 0$, $\hat{{\bf z}}$ is shifted further away from the origin in the latent space than ${\bf z}$.
Conversely, when $\xi' < 0$, $\hat{{\bf z}}$ is shifted closer 	to the origin.
We used the Glow model trained on the CIFAR-10’s training dataset and tested 1024 samples randomly selected from the test dataset of CIFAR-10, which thus corresponds to the In-Dist examples.
As shown in Fig.\ \ref{fig:xi_exp} (left), the degree of the reconstruction error is proportional to the intensity of the penalty, regardless of the sign of $\xi'$.
As a case of OOD, we also measured the same for the test samples of Celeb A, and similar results were obtained (Fig.\ \ref{fig:xi_exp} (right)).

\section{Experimental Setup}

\subsection{OOD datasets}
\label{sec:ood_datasets}
\subsubsection{Different datasets from In-Dist}
We use CelebA \cite{liu2015faceattributes}, TinyImageNet (TIN) \cite{ILSVRC15}, and LSUN \cite{journals/corr/YuZSSX15}
as OOD datasets.
The CelebA dataset contains various face images, from each of which we cut out a $148 \times 148$ centered area and re-scale them to the size of each In-Dist dataset.
The TIN is also used as OOD only when the In-Dist is C-10, re-scaling its images to $32 \times 32$.
For LSUN, we cut out squares with a length that matches the shorter of the height or width of the original image, followed by re-scaling to $32 \times 32$ for C-10 and $64 \times 64$ for TIN.

\subsubsection{Noise images} We control the noise complexity by following \cite{Serra2020Input}.
Specifically, noise images with the same size as In-Dist are sampled from uniform random, average-pooling is applied to them, and then they are resized back to the original size. The complexity is controlled by the pooling size, $\kappa$.
The images become most complex when $\kappa=1$ (i.e., no pooling) and become simpler as $\kappa$ is increased.
Treating images with different $\kappa$ as separate datasets, we refer to them as Noise-$\kappa$.

\subsection{Classifiers}
\label{sec:classifies}
	\begin{table*}[tb]
	\centering
	\caption{Classification accuracies (\%) for  adversarial examples.
		Results for \emph{no attack} are evaluated with the whole test dataset for each.
		Results for attacks are evaluated with 1024 adversarial examples we generated based on 1024 normal samples chosen at random from each test dataset.}
	\label{tab:acc_classifier}
	\vspace*{3mm}

	\begin{adjustbox}{max width=\textwidth}
		\begin{threeparttable}[t]
			\vspace{0.20cm}
			\scalebox{0.85}{
				\begin{tabular}{ccrrrrr}
					\toprule
					Dataset & Classifier &no attack & PGD-2 &  PGD-8  & CW-0 & CW-10   \\
					\midrule
					CIFAR-10 & WRN-28-10& 95.949 & 5.56 & 0.0 & 0.0 & 0.0 \\
					TinyImageNet  & WRN-28-10 & 66.450 & 2.92 & 1.95& 0.0 & 0.0 \\
					ILSVRC2012 &  ResNet-50 v2 & 67.628& 0.0 & 0.0 & 0.0 & 0.0 \\
					\bottomrule
				\end{tabular}
			}
		\end{threeparttable}
	\end{adjustbox}
\end{table*}
The archtecture of classifier we used for the experiments on CIFAR-10 and TinyImageNet  is WideResNet 28-10 (WRN 28-10) \cite{BMVC2016_87},
and the one on ILSVRC2012 is ResNet-50 v2 \cite{7780459}.
These classifiers are used for generating adversarial examples and also for the classifier-based comparison methods as described in Section \ref{sec:comparisons}.
The classification accuracies are 95.949 \%, 66.450 \%,  and 67.628 \% on the test datasets of C-10, TIN, ILSVRC2012, respectively (Table \ref{tab:acc_classifier}).

\subsection{Adversarial Examples}
\label{sec:adv_appendix}

\subsubsection{Attack methods}
\label{sec:attack_methods}
Let $C(\cdot)$ be a DNN classifier where $C_{i}({\bf x})$ denotes logit of class label $i$ for an image ${\bf x} \in [0,1]^d$, represented in the $d$ dimension normalized [0,1] space.
The predicted label is given as $y_{\text{pred}} = \argmax_{i} C_{i}({\bf x})$.
Using a targeted classifier $C$, we mount untargeted attacks with the two methods, that is, we attempt to generate adversarial examples ${\bf x}_{\text{adv}}$ to be labeled as $y_{\text{target}} = \argmax_{i \neq y_{\text{true}}} C_{i}({\bf x})$, where $y_{\text{true}}$ is the original label for ${\bf x}$.
In PGD, ${\bf x}_{\text{adv}}$ is searched within a hyper-cube around ${\bf x}$ with an edge length of $2 \epsilon$, which is written as
\begin{eqnarray}
	{\bf x}_{\text{adv}} = \min_{{\bf x}^{*}} \ L_{\text{pgd}}({\bf x}^{*}) \ \text{s.t.} \ \norm{{\bf x}^{*} - {\bf x}}_{\infty} \leq \epsilon,
	L_{\text{pgd}}({\bf x}^{*}) := C_{y_{\text{true}}}({\bf x}^{*}) - C_{y_{\text{target}}}({\bf x}^{*})
\end{eqnarray}
and $\epsilon$ is given as a hyper-parameter.
The CW attack is formalized as
\begin{eqnarray}
	{\bf x}_{\text{adv}}  = \min_{{\bf x}^{*}} \ \lambda L_{\text{cw}}({\bf x}^{*}) + \norm{{\bf x}^{*} - {\bf x}}_{2}^2,
	L_{\text{cw}}({\bf x}^{*}) := \max(C_{y_{\text{target}}}({\bf x}^{*}) - C_{y_{\text{true}}}({\bf x}^{*}), -k),
\end{eqnarray}
$k$ is a hyper-parameter called \emph{confidence}, and $\lambda$ is a learnable coefficient.
Defining a vector of adversarial perturbation $ \Delta_{\bf{x}} := {\bf x}_{\text{adv}} - {\bf x}$,
$\norm{\Delta_{\bf{x}}}_{\infty}$ in PGD is bounded by $\epsilon$, whereas $\norm{\Delta_{\bf{x}}}_{2}$ in CW is not.
Thus, while the artifacts may appear on images when $k$ becomes larger, CW attack always reaches 100\% successes in principle.

\subsubsection{Generation of adversarial examples}
\label{sec:gen_adv}
For PGD attacks, the step size is $\frac{1}{255}$ and the mini-batch size is 128 on all three datasets.
The numbers of projection iteration for PGD-2 are 1000 on C-10 and TIN and 100 on ILSVRC2012,
and those for PGD-8 are 100 on all three datasets.
For CW attacks, our code is based on the one used in \cite{carlini2017towards}.
The maximum numbers of iterations are 10000 on C-10 and 1000 on TIN and ILSVRC2012.
The number of times we perform binary search to find the optimal tradeoff-constant between $L_{2}$ distance and confidence is set to 10.
The initial tradeoff-constant to use to tune the relative importance of $L_{2}$  distance and confidence is set to 1e-3.
The learning rate for attacking is 1e-2 and the mini-batch size is 256 on C-10, 128 on TIN, and 64 on ILSVRC2012.
The classification accuracies before and after attacking are shown in Table \ref{tab:acc_classifier}.

\subsection{Glow}
\label{sec:glow_detail}
\paragraph{Architecture.}
The architecture of the Glow mainly consists of two parameters: the depth of flow $K$ and the number of levels $L$.
The set of affine coupling and $1 \times 1$ convolution are performed $K$ times, followed by the factoring-out operation \cite{dinh2016density}, and this sequence is repeated $L$ times.
We set $K=32$ and $L=3$ for C-10, $K=48$ and $L=4$ for TIN, and  $K=64$ and $L=5$ for ILSVRC2012.
Refer to \cite{NIPS2018_8224, dinh2016density} and our experimental code for more details.\footnote{We implemented Glow model based on \cite{code_Glow}.}
Many flow models including Glow employs \emph{affine coupling layer} \cite{dinh2016density} to implement $f_{i}$.
Splitting the input ${\bf x}$ into two halves as $[{\bf x}_{a}, {\bf x}_{b}] = {\bf x}$, a coupling is performed as $[{\bf h}_{a}, {\bf h}_{b}] = f_{i}({\bf x}) = [{\bf x}_{a}, {\bf x}_{b} \odot s({\bf x}_{a})+t({\bf x}_{a})]$ where $\odot$ denotes the element-wise product and $s$ and $t$ are the convolutional neural networks optimized through the training.
Its inverse can be easily computed as ${\bf x}_{a} = {\bf h}_{a}$ and ${\bf x}_{b} = ({\bf h}_{b} - t({\bf x}_{a})) \oslash s({\bf x}_{a})$, where $\oslash$ denotes the element-wise division, and $\text{log} \ |\text{det} \ J_{{f}_{i}}({\bf x})|$ can be obtained as just $\text{log} \  |s({\bf x}_{a})|$.
\paragraph{Restricted  affine scaling.}
In order to suppress the Lipschitz constants of the transformations and further improve stability,
we restrict the affine scaling to $(0.5, 1)$, following \cite{pmlr-v130-behrmann21a}.
Specifically, we replace $s({\bf x}_{a})$ with $g(s({\bf x}_{a}))$ and use a half-sigmoid function as $g(s) = \sigma(s)/2 + 0.5$ where $\sigma$ is the sigmoid function.
\paragraph{Training settings.}
We trained $45100$ iterations for C-10 and TIN and  $70100$ iterations for ImaneNet, using the Adam optimizer \cite{kingma2014adam} with $\beta_{1} = 0.9$ and $\beta_{2} = 0.999$. The batch size is $256$ for C-10, $16$ for TIN, and $4$ for ILSVRC2012.
The learning rate for the first $5100$  iterations is set to 1e-4, and 5e-5 for the rest.
We applied the data augmentation of random $2 \times 2$ translation and horizontal flip on C-10 and TIN.
For ILSVRC2012, 
the image is first resized to be $256$ in height or width, whichever is shorter, and cropped to $224 \times 224$ at random.

\subsection{Existing Methods}
\label{sec:comparisons_detail}
We compare the performance of the proposed methods to several existing methods.
In the following description for the likelihood-based methods (i.e., WAIC, LLR, and COMP), $p({\bf x})$ denotes the likelihood computed with the Glow model, the same as the one we use for our methods.
We reverse the sign of score outputted from those likelihood-based methods since we use them as scores indicating being OOD examples.
For the classifier-based methods (i.e., DU, FS, and PL), the classifier ($C({\bf x})$) is the WRN 28-10 or ResNet-50 v2 which is the same model we used to craft the adversarial examples in the previous section.
The comparison methods are as follows:

\setlength{\leftmargini}{0.5cm}
\begin{enumerate}

	\item The Watanabe-Akaike Information Criterion (WAIC) \cite{choi2019generative} measures $\mathbb{E}[\text{log} \ p({\bf x})] - \text{Var}[\text{log} \ p({\bf x})]$ with an ensemble of five Glow models.
	The four of the five were trained separately with different affine-scaling restrictions mentioned in Section \ref{sec:glow}, i.e., the function $g(s)$ described in Appendix \ref{sec:glow_detail}.
	The function $g(s)$ we chose for the four models is the following: a sigmoid function $\sigma(s)$, a half-sigmoid function $\sigma(s)/2 + 0.5$, clipping as ${\tt min}(|s|, 15)$, and an additive conversion as $g(s) = 1$.
	In addition to those four models, we used the background model used in the LLR described next as a component of the ensemble.

	\item The likelihood-ratio test (LLR) \cite{NEURIPS2019_1e795968} measures $\text{log} \ p({\bf x}) - \text{log} \ p_0({\bf x})$. The $p_0({\bf x})$  is a background model trained using the training data with additional random noise sampled from the Bernoulli distribution with a probability 0.15. It thus uses two Glow models.

	\item The Complexity-aware likelihood test (COMP) \cite{Serra2020Input} measures $\text{log} \ p({\bf x}) + \frac{|B({\bf x})|}{d}$ where $B({\bf x})$ is a lossless compressor that outputs a bit string of ${\bf x}$ and its length $|B({\bf x})|$ is normalized by $d$, the dimensionality of ${\bf x}$. We use a PNG compressor as $B$.

	\item The typicality test in latent space (TTL) \cite{choi2019generative} measures the Euclidean distance in the  latent space to the closest point on the Gaussian Annulus:
	${\tt abs} (\norm{ f({\bf x})} - \sqrt{d})$ where $f$ is the same Glow model used in our method.

	\item The Maximum Softmax Probability (MSP) \cite{hendrycks2017a} simply measures $ \tt{max} ( {\tt softmax}(C({\bf x}))$, which has been often used as a baseline method in the previous OOD detection works.

	\item The Dropout Uncertainty (DU) \cite{feinman2017detecting} is measured by Monte Carlo Dropout \cite{gal2016dropout}, i.e., the sum of the variance of each component of ${\tt softmax}(C({\bf x}))$, computed over 30 times run with the dropout rate 0.2.

	\item The Feature Squeezing (FS) \cite{FS_ndss18_osada} outputs $L_{1}$ norm distance as the score: $\norm{{\tt softmax}(C({\bf x})) - {\tt softmax}(C(\hat{{\bf x}}))}_{1}$ where $\hat{{\bf x}}$ is generated by applying a median filter to the original input ${\bf x}$.

	\item The Pairwise Log-Odds (PL) \cite{pmlr-v97-roth19a} measures how logits change under random noise. Defining perturbed log-odds between classes $i$ and $j$ as $G_{i j}({\bf x}, \epsilon) := C_{i j}({\bf x} + \epsilon) - C_{i j}({\bf x})$ where $C_{i j}({\bf x}) := C_{j}({\bf x}) - C_{i}({\bf x})$, the score is defined as
	$\tt{max}_{i \neq y_{\text{pred}}} \mathbb{E}[G_{i y_{\text{pred}}}({\bf x},\epsilon)]$
	where $y_{\text{pred}} = \argmax_{i} C_{i}({\bf x})$.
	The noise $\epsilon$ is sampled from $\text{Uniform}(0,1)$, and we took the expectation over 30 times run.

	\item The reconstruction error in Auto-Encoder is defined as $\text{AE}({\bf x}) = \norm{{\bf x} - f_{\text{d}} (f_{\text{e}}({\bf x})))}_{2}$ where $f_{\text{e}}$ and $f_{\text{d}}$ are the encoder and decoder networks, respectively. The architecture of the AE model we used and its training procedure is presented in Appendix \ref{sec:ae}.
\end{enumerate}

\subsubsection{Auto-Encoder}
\label{sec:ae}
\begin{table*}[h]
	\caption{Architecture of encoder and decoder of Auto-Encoder. BNorm stands for batch normalization. Slopes of Leaky ReLU (lReLU) are set to $0.1$.}
	\centering
		\begin{tabular}{ll}
			\toprule
			Encoder & Decoder \\
			\midrule
			Input: $32 \times 32 \times 3$ image & Input: 256-dimensional vector\\
			$3 \times 3$ conv. 128 same padding, BNorm, ReLU \ & Fully connected 256 $\rightarrow$ 512 ($ 4 \times 4 \times 32$), lReLU \\
			$3 \times 3$ conv. 256 same padding, BNorm, ReLU \ & $3 \times 3$ deconv. 512 same padding, BNorm, ReLU \\
			$3 \times 3$ conv. 512 same padding, BNorm, ReLU   & $3 \times 3$ deconv. 256 same padding, BNorm, ReLU \\
			Fully connected 8192 $\rightarrow$ 256     & $3 \times 3$ deconv. 128 same padding, BNorm, ReLU \\
			& $1 \times 1$ conv. 128 valid padding, sigmoid \\
			\bottomrule
		\end{tabular}
	\label{tab:ae}
\end{table*}
We use the Auto-Encoder based method as one of the comparing methods in our experiments on CIFAR-10 and TinyImageNet.
The architecture of the Auto-Encoder we used is designed based on DCGAN \cite{radford2015unsupervised}, which is shown in Table \ref{tab:ae}.
For both datasets, we used the same architecture and applied the following settings.
We used the Adam optimizer with $\beta_{1} = 0.9$ and $\beta_{2} = 0.999$, with batch size $256$.
The learning rate starts with $0.01$ and exponentially decays with rate $0.8$ at every $2$ epochs, and we trained for $300$ epochs.
The same data augmentation as used for Glow was applied.

\section{More Experimental Results}

\subsection{AUPR}
\label{sec:aupr}
We show  the detection performance as the area under the precision-recall curve (AUPR).
Table \ref{tab:aupr_C10} shows the results on CIFAR-10, Table \ref{tab:aupr_TIN} shows the results on TinyImageNet, and  Table \ref{tab:aupr_IN} shows the results on ILSVRC2012.

\begin{table*}[h]
	\caption{AUPR (\%) on CIFAR-10. The column labeled as \bsq{Avg.} shows the averaged scores. }
	\label{tab:aupr_C10}
	\centering
	\begin{adjustbox}{max width=\textwidth}
		\begin{threeparttable}[t]
			\vspace{0.20cm}
			\scalebox{0.85}{
				\begin{tabular}{crrrrrrrrrrrr}
					\toprule
					              & CelebA    & TIN  & Bed & Living & Tower  &  PGD-2  & PGD-8 & CW-0 & CW-10 & Noise-1 & Noise-2  &  \underline{Avg.} \\
					\midrule
					WAIC    & 50.24 & 74.29 & 80.70 & 86.54 & 83.69  & 43.88 & 70.07 & 48.26 & 47.20 & {\bf 100.0} & {\bf 100.0}& 71.35 \\
					LLR        & 62.59  & 43.00 & 44.81 & 43.81 & 39.83 & 57.83  & 78.54 & 51.87 & 54.06 & 30.82 & 30.71 &  48.90 \\
					COMP   & 75.37 & 74.88 & 73.32 & 84.95 & 58.80 &  61.18 & 98.49 & 50.95 & 53.68 & {\bf 100.0} & {\bf 100.0}& 75.60 \\
					TTL        & 82.49 & 85.58 & 90.68 & 91.60 & 89.88 & 73.82 & {\bf 99.99} & 50.57 & 53.90 & {\bf 100.0}  & 51.04 &  79.05 \\
					MSP     & 78.16 & 88.74 & {\bf 94.58} & 92.19 & {\bf 92.74} & 35.79 & 30.69 & {\bf 99.15} & 31.63 &  98.82 & 97.40 &  76.35\\
					PL      & 76.84 & 57.80 & 51.59 & 47.01 & 51.40 & 79.88  & 96.98 & 53.00  & 81.00  & 35.53 & 65.84 &  63.35 \\
					FS      & 76.96  & 82.63 & 80.41 & 80.22& 78.74 & {\bf 91.51} & 75.31 & 83.80 & {\bf 94.84 }& 81.69 & 90.70 & 83.37  \\
					DU      & 76.86 & 75.43 & 75.79 & 74.42 & 74.78 & 70.52 & 36.20 & 85.63 & 76.44 & 69.36 & 65.49 &  70.99 \\
					AE      & 59.61 & 81.63 & 70.36 & 84.71 & 66.37& 50.37 & 56.71 & 50.04 &49.96  & 98.74 &   {\bf 100.0} & 69.86 \\
					\midrule
					RE (ours)  & 86.06 & 87.27 & 88.33 & 87.99 & 87.98 & 84.23  & 87.44 & 90.05 & 89.89 & 90.78 & 89.16 &  88.11 \\
					PRE (ours) & {\bf 88.42} & {\bf 92.68} & 92.12 & {\bf 92.23} & 92.32  & 84.78  & {\bf 99.99} & 89.98 & 90.09 & {\bf 100.0}& 88.91 &  {\bf 91.96} \\
					\bottomrule
				\end{tabular}
			}
		\end{threeparttable}
	\end{adjustbox}
\end{table*}

\begin{table*}[h]
	\caption{AUPR (\%) on TinyImageNet. The column labeled as \bsq{Avg.} shows the averaged scores.
	}
	\label{tab:aupr_TIN}
	\centering
	\begin{adjustbox}{max width=\textwidth}
		\begin{threeparttable}[t]
			\vspace{0.20cm}
			\scalebox{0.85}{
				\begin{tabular}{crrrrrrrrrrr}
					\toprule
					& CelebA     & Bed & Living & Tower &   PGD-2  & PGD-8 & CW-0 & CW-10 & Noise-1 & Noise-2 &  \underline{Avg.} \\
					\midrule
					WAIC    & 32.54 & 59.41 & 63.02 & 63.81  & 42.89 & 44.77& 48.84 & 47.00 & {\bf 100.0} & {\bf 100.0} &   60.23 \\
					LLR       & 90.81 & 58.27 & 58.43 & 70.46  & 56.81 & 94.75 & 51.35 & 53.37 & 70.30 & 30.69 &  63.52  \\
					COMP  & 50.31 & 50.92 & 62.30 & 46.43  & 56.49 & 94.86 & 50.75 & 52.01 & {\bf 100.0} & {\bf 100.0}  & 66.41  \\
					TTL  	  & {\bf 96.27} & {\bf 99.03} & {\bf 99.45} & {\bf 98.90} & 78.61 & {\bf 100.0} & 50.96 & 55.19 & {\bf 100.0} & {\bf 100.0}  &    87.84  \\
					MSP     & 79.60 & 81.29 & 77.42 & 76.76 & 31.44 & 30.69 & 76.13 & 30.93 &   79.53 &  83.20 & 64.70 \\
					PL         & 46.71 & 34.51 & 39.27 & 38.11  & {\bf 93.26} & 99.99& 41.59 & {\bf 95.95} & 33.40 & 39.33 &  56.21   \\
					FS         & 36.83 & 36.23 & 36.90 & 35.77  &78.27  & 39.19 & 46.41 & 86.10 & 41.45 & 45.52 &  48.27  \\
					DU        & 44.73 & 41.07 & 38.91 & 37.41 & 50.20 & 68.04  & 48.88 & 49.82 & 44.56 & 34.93 &  45.86  \\
					AE         & 33.24 & 34.50 & 38.11 & 35.24   & 49.64 & 50.63 & 49.98 & 49.95 & 85.97 & 56.33 &  48.86  \\
					\midrule
					RE (ours)  &46.42  & 56.06 & 56.28 & 57.51   & 87.98 & 88.71  & {\bf 92.04} & 93.10 & 94.12 & 93.81 &  76.60 \\
					PRE (ours) & 70.62 & 92.51 & 94.96 & 91.81  & 88.41 & 99.90 & 91.83 & 92.90 & {\bf 100.0} & {\bf 100.0} &   {\bf 92.29}\\
					\bottomrule
				\end{tabular}
			}
		\end{threeparttable}
	\end{adjustbox}
\end{table*}

\begin{table*}[h]
	\caption{AUPR (\%) on ILSVRC2012 with our method. The column labeled as \bsq{Avg.} shows the averaged scores.}
	\label{tab:aupr_IN}
	\centering
	\begin{adjustbox}{max width=\textwidth}
		\begin{threeparttable}[t]
			\vspace{0.20cm}
			\scalebox{0.85}{
				\begin{tabular}{crrrrrrrr}
					\toprule
					& CelebA     & PGD-2     & PGD-8 & CW-0 & CW-10 & Noise-2 & Noise-32 &  \underline{Avg.} \\
					\midrule
					RE   & {\bf  90.04} & 88.59 & 90.35 & {\bf 90.52} & 91.08 & {\bf 91.42} & 91.45  &  90.49 \\
					PRE  &{\bf  90.04} & {\bf 88.60} & {\bf 90.47} & {\bf 90.52} & {\bf 91.09} & {\bf 91.42} & {\bf 91.46}  &  {\bf 90.51}  \\
					\bottomrule
				\end{tabular}
			}
		\end{threeparttable}
	\end{adjustbox}
\end{table*}


\subsection{Effects of Coefficient of Penalty}
\label{sec:lambda}
Table \ref{tab:lambda} shows the AUROC with different $\lambda$ in Eq.\ \eqref{eq:rxi}, the coefficients of $\xi$.
We empirically chose $\lambda=50$ for CIFAR-10 and $\lambda=100$ for the other two datasets based on these results.

\begin{table}[h]
	\caption{AUROC (\%) with various coefficient of penalty, $\lambda$. }
	\label{tab:lambda}
	\centering
	\begin{adjustbox}{max width=\textwidth}
		\begin{threeparttable}[t]
			\vspace{0.20cm}
			\scalebox{0.85}{
				\begin{tabular}{rrrrrrr}
					\toprule
					In-Dist & \multicolumn{2}{c}{CIFAR-10}  & \multicolumn{2}{c}{TinyImageNet}  & \multicolumn{2}{c}{ILSVRC2012}                   \\
					\cmidrule(r){2-3} 	\cmidrule(r){4-5}  	\cmidrule(r){6-7}
					$\lambda$ / OOD & CelebA   & PGD-2  & CelebA   & PGD-2   & CelebA   & PGD-2   \\
					\midrule
					0  & 92.53  & 91.66  & 46.68  & 92.86  & 94.65  & 93.96 \\
					10      & 93.40  & {\bf  92.95}  & 79.99  & 94.05  & 94.64  & 93.96 \\
					50     & {\bf 93.62}  & 92.23  & 93.13  & {\bf 95.13}  & 94.56  & 93.93 \\
					100   & 93.60  & 92.09  & 95.55  & 95.04  & {\bf  94.89} & {\bf  94.24} \\
					500   & 91.94  & 89.21  & {\bf  98.41}  & 92.53  & 93.02  & 92.36 \\
					1000 & 90.81  & 85.08  & 98.33  & 90.83  & 88.92  & 89.30 \\
					\bottomrule
				\end{tabular}
			}
		\end{threeparttable}
	\end{adjustbox}
\end{table}

\subsection{Histograms of $L_{2}$ norm for partitioned latent vector}
\label{sec:za_zb}
\begin{figure}[h]
	\centering
	\includegraphics[width=0.22 \textwidth]{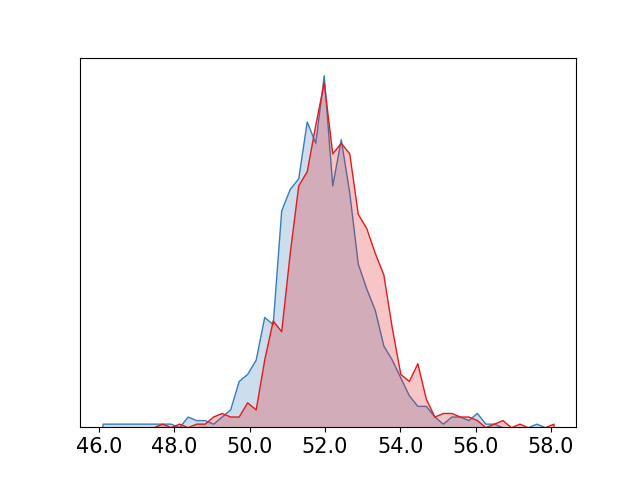}
	\includegraphics[width=0.22 \textwidth]{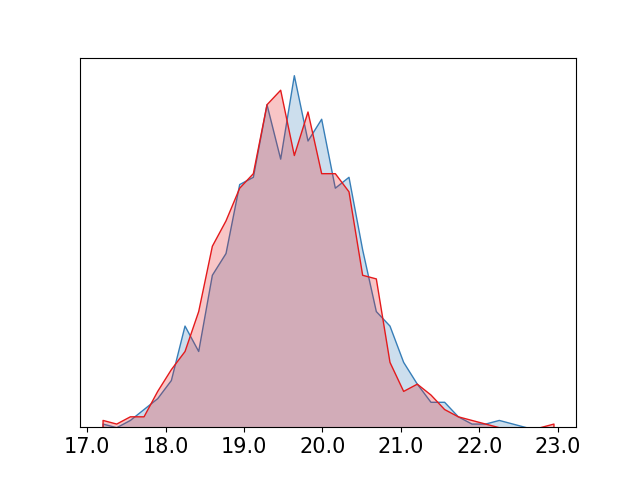}
	\caption{Histograms for $L_{2}$ norm for ${\bf z}_{a} \in \mathbb{R}^{2688}$ (left) and ${\bf z}_{b} \in \mathbb{R}^{384}$ (right). Red is the distribution for the OOD: CW-0's examples on CIFAR-10, and blue is In-Dist's examples.
	}
	\label{fig:za_zb}
\end{figure}

The typicality test (TTL) failed to detect CW’s adversarial examples (and some Noise datasets), while the PRE and RE successfully did, as shown in Tables \ref{tab:auroc_C10} and \ref{tab:auroc_TIN}.
The failure of TTL is obvious from Fig.\ \ref{fig:hist_dz} showing that the distributions of $\norm{{\bf z}}$ for the CW’s examples (the adversarial examples generated by CW-0) overlap those for In-Dist’s almost entirely.
We analyzed this failure by the following procedure.
We partitioned latent vectors ${\bf z} \in \mathbb{R}^{3072}$ on CIFAR-10 into two parts as [${\bf z}_{a}, {\bf z}_{b}] = {\bf z}$  where ${\bf z}_{a} \in \mathbb{R}^{2688}$ and ${\bf z}_{b} \in \mathbb{R}^{384}$, and measuring the $L_{2}$ norm separately for each.
\footnote{This partitioning is based on the factoring-out operation \cite{dinh2016density} used in affine coupling-type NFs, including Glow. The inputs with 3072 dimension are partitioned this way when factoring-out is applied three times.}
In Fig.\ \ref{fig:za_zb}, we show the distributions comparing the CW-0's example with the In-Dist's  in $\norm{{\bf z}_{a}}$ and $\norm{{\bf z}_{b}}$.
The distribution of $\norm{{\bf z}_{a}}$ for the CW-0’s examples is shifted toward larger values than that for In-Dist’s.
In contrast, the opposite is true for $\norm{{\bf z}_{b}}$, where the distribution for the CW-0’s examples is slightly shifted toward smaller values than that for In-Dist’s.
When we measure the $L_{2}$ norm for the entire dimension of ${\bf z}$ without partitioning, the deviation from the In-Dist observed in ${\bf z}_{a}$ and ${\bf z}_{b}$ cancels out, and as a result, $\norm{{\bf z}}$ of CW's examples becomes indistinguishable from those of In-Dist ones.
This is exactly the case described in Section \ref{sec:tt} where the typicality test fails.

\subsection{Histograms for Reconstruction Error without Penalty, $R$}
\label{sec:dx}
\begin{figure}[h]
	\centering
	\includegraphics[width=0.32 \textwidth]{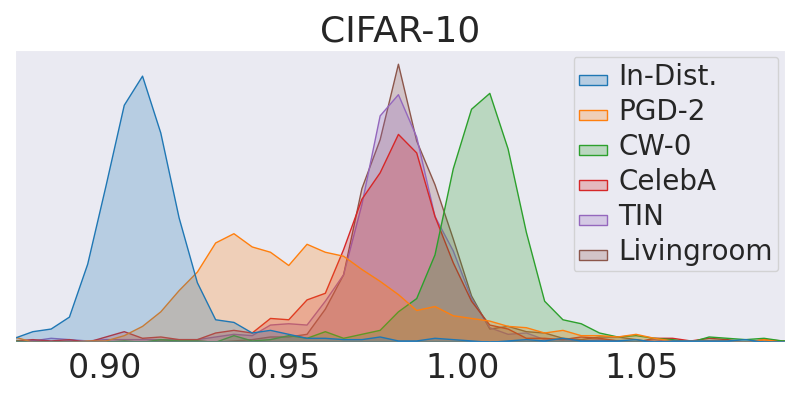}
	\includegraphics[width=0.32 \textwidth]{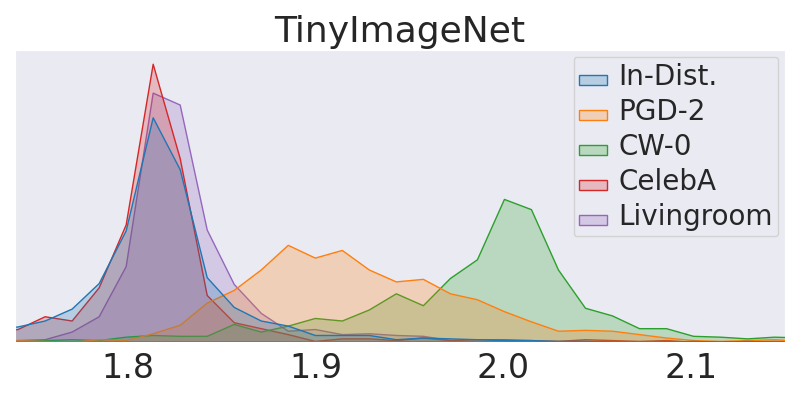}
	\includegraphics[width=0.32 \textwidth]{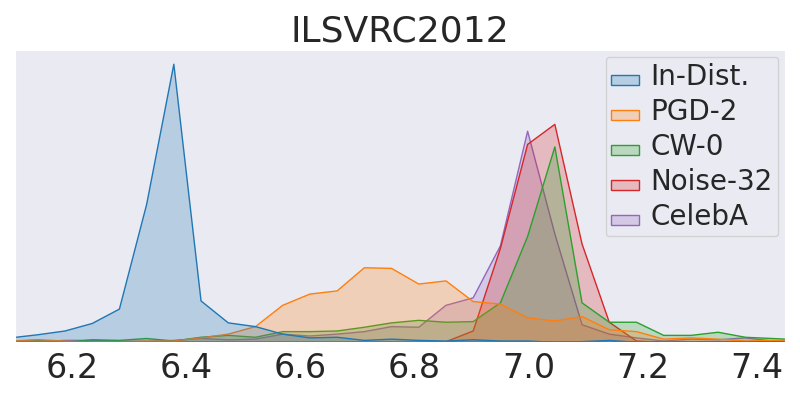}
	\caption{Histograms for reconstruction error without penalty in the latent space. The x-axis is $R$.
	}
	\label{fig:hist_dx}
\end{figure}

Fig.\ \ref{fig:hist_dx} shows the histograms of reconstruction error without penalty in the latent space, $R$ in correspondence with Fig.\ \ref{fig:hist_dx_ga}.

\section{Analysis with Chernoff Tail Bound}
\label{sec:tailbound}
Let ${\bf z} \in \mathbb{R}^{d}$ be a random vector sampled from a standard Gaussian.
Then, the following is obtained from Markov's inequality:
\begin{eqnarray}
	\text{Pr} \left[ \norm{{\bf z}}_{2}  > \sqrt{d(1 + \epsilon)} \right]  \leq   \text{exp} \left(- \frac{d \epsilon^{2}}{8} \right), \\
	\text{Pr} \left[ \norm{{\bf z}}_{2}  < \sqrt{d(1 - \epsilon)} \right]  \leq   \text{exp} \left(- \frac{d \epsilon^{2}}{8} \right).
\end{eqnarray}
We refer the reader to \cite{osadaChernoffTailBound}, for example, for the derivation.
From the above, we obtain \eqref{eq:tailbound}.
Letting $\text{exp} \left(- \frac{d  \epsilon^{2}}{8} \right) = \frac{1 }{ 2^{s}}$, we have 
$ \epsilon = \sqrt{\frac{8s}{d \text{log}_{2}(e)}} \approx \sqrt{\frac{8s}{d \cdot 1.4427}}$.
By setting $s=58$ (and $d=3072$) which results in $ \epsilon= 0.32356413$, the inequality is obtained.
\end{document}